\documentclass[10pt,twocolumn,letterpaper]{article}

\usepackage{cvpr} 
\usepackage{times}
\usepackage{epsfig}
\usepackage{graphicx}
\usepackage{amsmath}
\usepackage{amssymb}

\usepackage{rotating}
\usepackage{booktabs}
\usepackage{multirow}
\usepackage[normalem]{ulem}

\usepackage{adjustbox}

\usepackage{mdframed}

\usepackage{caption}
\usepackage{cuted}
\usepackage{float}

\usepackage[pagebackref=true,breaklinks=true,letterpaper=true,colorlinks,bookmarks=false]{hyperref}

\begin{document}

\title{Beyond Error-vs-Discard Characteristic: Toward Stable and Reliable Evaluation for Face Image Quality Assessment}



\author{
Bhavesh Wani$^{1}$ \quad
Žiga Babnik$^{2}$ \quad
Vitomir Štruc$^{2}$ \quad
Philipp Terhörst$^{1}$\\
$^{1}$Johannes Gutenberg University Mainz, Germany \quad
$^{2}$University of Ljubljana, Slovenia
}

\maketitle
\thispagestyle{empty}

\begin{abstract}
Face Image Quality Assessment (FIQA) aims to estimate the utility of facial images for reliable recognition. The evaluation of FIQA methods is predominantly based on the Error-versus-Discard Characteristic (EDC), which evaluates performance by progressively discarding low-quality samples and measuring recognition error on the retained subset. In this work, we demonstrate that the widely used EDC protocol has fundamental limitations: Test-Set Divergence and Threshold Drift, which together limit the reliability and comparability of FIQA methods. To address this, we propose discard-based EDC variants and a rank-based Rank Consistency Evaluation (RCE) metric that operates on the entire test set without discarding samples, using a fixed decision threshold. Extensive experiments on five datasets, four face recognition models, and 15 state-of-the-art FIQA methods demonstrate both the limitations of EDC and the effectiveness of the proposed approaches in enabling a more reliable and comparable evaluation.
Despite evaluated on face images only, the limitations arise from the EDC protocol rather than the biometric modality, suggesting a broader applicability to biometric quality assessment in general.
Code: \url{https://github.com/RAIB-group/RCE}

\end{abstract}

\section{Introduction}

Face Image Quality Assessment (FIQA) plays a crucial role in biometric systems, as the quality of face images directly affects the system performance, bias, and reliability~\cite{ISO29794-1-2016,4107559,terhorst2020face,6095497}. In practical applications, FIQA is commonly used to discard low-quality samples that are likely to cause verification errors during deployment. Moreover, quality estimates can be integrated in the decision-making process to improve performance~\cite{Meng2021MagFaceAU,Kim2022AdaFaceQA,qmagface}. Since FIQA significantly affects the system, reliable evaluation of FIQA methods is essential for real-world deployment.


The Error-versus-Discard Characteristic (EDC) is the de facto standard protocol for evaluating FIQA methods, where samples (images) are ranked by predicted quality and low-quality samples are progressively removed~\cite{ISO29794-1-2016,4107559,6095497}. The recognition error is measured on the remaining data, typically using the false non-match rate (FNMR) and false match rate (FMR), and a FIQA method is considered effective if discarding potentially poor-quality images quickly reduces the error rate. However, despite its widespread use, prior work identified notable limitations. Biometric samples lack a standardized ground truth and are assessed indirectly via recognition performance~\cite{4107559}. The protocol is also sensitive to parameters such as discard step size and interpolation, while dataset and experimental variations hinder comparability across FIQA methods~\cite{10330743_Considerations}.


This study identifies two key limitations of EDC-based evaluation that undermine the reliability and comparability of FIQA methods.
(1) During EDC computation, recognition performance is evaluated on progressively different datasets due to successive discards, leading to a steady decline in comparability, which we term \textit{Test-Set Divergence}.
(2) Additionally, recognition performance is computed not only on different datasets but with a decision threshold that is recomputed at each discard step, resulting in various \textit{Threshold Drift}s, where FIQA methods are compared at substantially different operating points, further reducing comparability.
To address these limitations, we first propose discard-based EDC variants, including fixed-threshold (FT-EDC) and proportional pair discard (PPD-EDC), as well as their combination (FT-PPD-EDC). Furthermore, we introduce a Rank Consistency Evaluation (RCE) framework, which shifts from discard-based to rank-based evaluation by operating on the entire test set with a fixed decision threshold and directly evaluating the correlation between predicted quality and recognition errors to allow comparable evaluations.


We demonstrate the limitations of EDC-based evaluation and the effectiveness of the proposed alternatives in comprehensive experiments conducted across five datasets, four face recognition (FR) models, and 15 state-of-the-art FIQA methods.
This extensive analysis provides a more reliable, consistent, and comparable overview of current FIQA approaches and shows that the most recent methods, which perform well under EDC-based evaluation, often perform considerably worse under RCE-based evaluation.

\section{Background and Related Work}



Biometric sample quality evaluation estimates a sample’s recognition utility by predicting verification errors~\cite{ISO29794-1-2016,4107559,6095497}.
In 2007, Grother et al.~\cite{4107559} introduced Error-versus-Reject Curves (ERC), later formalized as the Error-versus-Discard Characteristic (EDC), now the de facto standard for biometric quality assessment and FIQA evaluation. EDC is widely used in academic research and large-scale benchmarks, including the NIST Face Recognition Vendor Test (FRVT)~\cite{fiqa_survey}.



In \cite{10330743_Considerations}, EDC-based evaluation is shown to be sensitive to discard step size, operating threshold, interpolation, and score normalization. FIQA method rankings vary across configurations: small discard steps reduce stability, while linear interpolation and inappropriate normalization may distort results. Utility-based evaluation~\cite{henniger2022utility,henniger2024utility} defines sample quality by its contribution to recognition performance, typically based on the separation between mated and non-mated score distributions. These approaches provide sample-wise quality estimates but are commonly evaluated using discard-based protocols.

Both EDC and utility-based approaches rely on discard-based evaluation, where the test set changes with the discard rate, limiting comparability across methods and demonstrating the need for more reliable evaluation methods.

\section{Limitations of the Error-versus-Discard Characteristic (EDC)}

\begin{figure}[t]
    \centering
    \includegraphics[
        width=0.799\linewidth,
        keepaspectratio
    ]{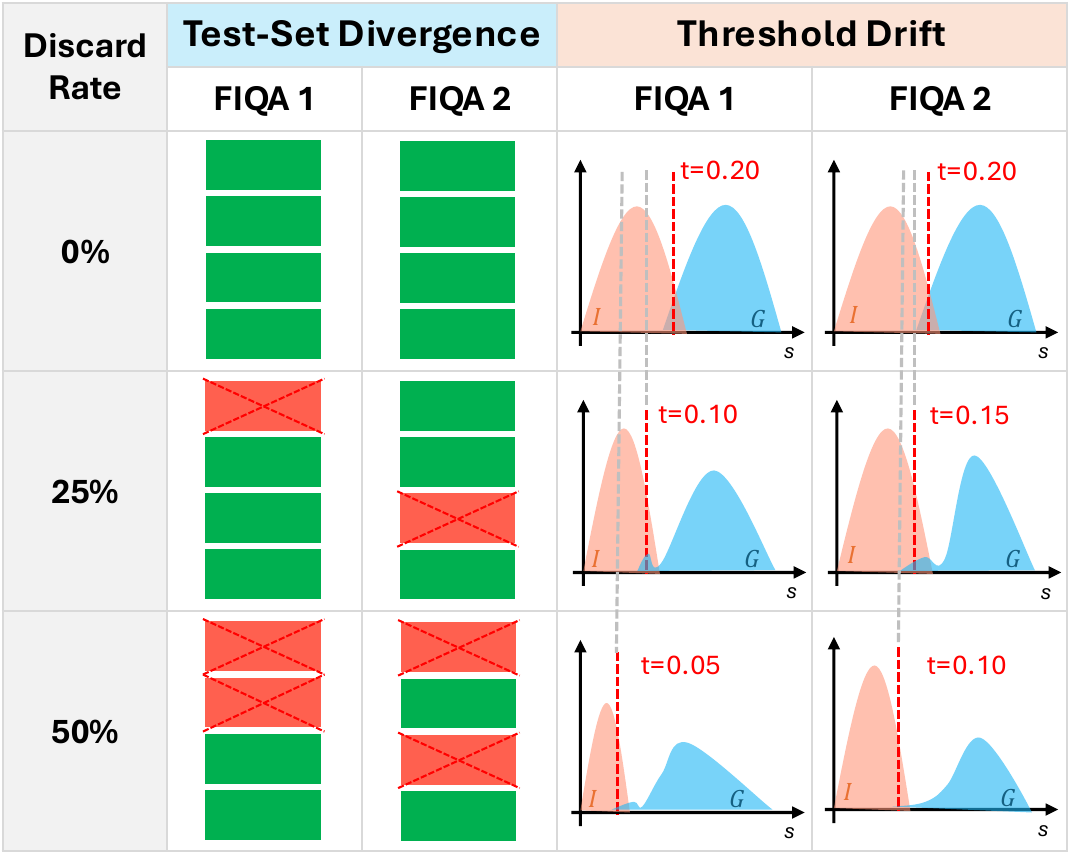}
    \caption{\textbf{Limitations of the EDC Protocol.} With increasing discard rates the FIQA methods (a) are evaluated on progressively different test sets (Test-Set Divergence) and (b) operate on completely different opertation points (Threshold Drift), leading to limited comparability of different FIQA methods.
    }
    \label{fig:edc_limitations}
    \vspace{-0.35cm}
\end{figure}

EDC is the de facto protocol for evaluating FIQA methods, but its discard-based process introduces key limitations. Different FIQA methods operate on distinct retained subsets, causing \textit{Test-Set Divergence}. The decision threshold is recomputed at each discard step to maintain the target FMR, resulting in \textit{Threshold Drift}. Together, these effects reduce evaluation stability and comparability (Fig.~\ref{fig:edc_limitations}).


\subsection{Test-Set Divergence}


In EDC, a FIQA method assigns a quality score to each sample and progressively removes the lowest-quality samples. Because different FIQA methods produce different scores, they retain and discard different subsets of samples. Even at the same discard rate, two methods eliminate different sample sets. As a result, the retained test sets are not directly comparable from the first discard step. This divergence grows with each subsequent step.


Moreover, discarding samples does not affect genuine and impostor pairs uniformly. A sample belonging to an identity with many images contributes to more genuine pairs than a sample from a sparsely represented identity.
Consequently, even when two low-quality samples have identical true quality and generate the same number of mismatches, removing the sample from the more populated identity eliminates far more genuine comparisons than removing the sample from the sparsely populated identity, leading to different EDC results.
Because FNMR is the proportion of genuine attempts incorrectly rejected by the system, discarding a sample that contributes many genuine pairs has a larger effect on the observed FNMR. In this sense, EDC implicitly rewards FIQA algorithms that assign low-quality scores to samples from under-represented identities, since their removal has little effect on the overall FNMR. An analogous argument applies to impostor pairs.
This implies that test-set characteristics, such as the genuine-to-impostor ratio, may differ across FIQA methods. As discarding proceeds, these differences in the retained test sets and their statistical properties grow, making FIQA methods increasingly difficult to compare.

\begin{mdframed}
\noindent
EDC’s sample‑dependent discarding process generates divergent test sets with differing genuine‑impostor ratios, which bias verification error rates and make FIQA methods progressively incomparable.
\end{mdframed}

\subsection{Threshold Drift}

In EDC, at each discard step, the decision threshold is recomputed to maintain a target $\mathrm{FMR}_{\text{target}}$. As samples are removed, the similarity score distributions of both genuine and impostor pairs change with the discard rate $r$, affecting the operating threshold, which is defined as
\begin{equation}
\tau(r) = \arg\min_{\tau} \left| FMR_r(\tau) - FMR_{\text{target}} \right|,
\end{equation}
where $\mathrm{FMR}_r(\tau)$ is computed at threshold $\tau$ using the retained impostor pairs $I_r$ at discard rate $r$.

As the number of retained impostor pairs $|I_r|$ decreases, fewer FMR values close to the target FMR can be computed. 
A lower threshold increases FMR by incorrectly accepting more impostor pairs, whereas a higher threshold decreases FMR but increases FNMR.
This leads to instability in the threshold computation, where small changes in the retained data can cause noticeable shifts in the threshold, leading to a noisy threshold selection. 


More importantly, despite threshold-selection instability, recomputing the threshold at each discard step leads to a \textit{Threshold Drift}, as the decision threshold is constantly drifting due to the shrinking comparison score distributions. As these score distributions change differently for every FIQA method, the drift is inconsistent across FIQA methods. 

\begin{mdframed}
\noindent
EDC’s per‑step threshold recomputation causes method‑specific threshold drift and instability due to shifting score distributions, making FIQA evaluations noisy and less comparable, as methods are compared at completely different operation points.
\end{mdframed}

\section{Methodology}

To address the limitations of EDC: \textit{Test-Set Divergence} and \textit{Threshold Drift}, we propose alternative evaluation strategies. 
These include EDC extensions to allow for a more gradual extension in the community without changing the entire framework and Rank Consistency Evaluation (RCE) method the eliminates both limitations completely.

\subsection{Fixed-Threshold EDC (FT-EDC)}
\label{sec:FT-EDC}

To eliminate the threshold‑drift problem inherent in the standard EDC protocol, we adopt a single global decision threshold, $\tau$, computed once on the full test set.  
The threshold is chosen so that the system operates at a predefined FMR.  
After $\tau$ is fixed, all subsequent discard steps reuse the same threshold: for each discard rate $r$, the lowest‑quality samples are removed, and the FNMR and FMR are recomputed using the fixed $\tau$. This guarantees that the decision boundary remains identical across all discard levels.

Because FNMR and FMR change jointly as $r$ varies, we summarise them with a single scalar metric, the Combined Error Rate (CER):
\begin{equation*}
\mathrm{CER}(r)=\lambda\;\mathrm{FMR}(r)+(1-\lambda)\;\mathrm{FNMR}(r),
\end{equation*}
where $\lambda\in[0,1]$ governs the relative importance of the two error types.  
In the present study, we set $\lambda=0.5$, assigning equal weight to false matches and false non‑matches, which follows the balanced evaluation used in~\cite{de2021fairness}.
While the FT-EDC scheme solves the \textit{Threshold Drift} intuitively, its biggest drawback is the need to choose the importance trade-off $\lambda$ based on the expected error rate magnitude and its intended application.

\subsection{Proportional Pair Discard EDC (PPD-EDC)}
\label{sec:PPD-EDC}
In EDC, progressive discard of low-quality samples alters the composition of genuine and impostor pairs. Discarding low-quality samples can remove a different proportion of genuine and impostor pairs, thereby affecting the class distribution of the retained set and the associated error computation for the EDC curve. To address this issue, a proportional pair discard (PPD) EDC strategy is proposed that ensures a constant genuine-to-imposter ratio.

More precisely, PPD-EDC replaces only the discard mechanism of the original EDC procedure. Let each image $i$ in the verification set be assigned a quality score $Q_i$. 
For a verification pair $(i,j)$, we define the pair quality as
$q_{ij}=\min\{Q_i,Q_j\}$.
All genuine pairs are collected in $\mathcal{G}$ and all impostor pairs in $\mathcal{I}$, with $|\mathcal{G}| = G$ and $|\mathcal{I}| = I$. Both $\mathcal{G}$ and $\mathcal{I}$ are sorted in ascending order of $q_{ij}$.
For a chosen discard rate $r \in [0,1]$, the number of pairs to be removed is $k = r \, (G + I)$.
The removal is performed proportionally to preserve the original genuine-to-impostor ratio:
\begin{equation}
k_G = \left\lfloor k \cdot \frac{G}{G + I} \right\rceil, \qquad
k_I = k - k_G,
\end{equation}
The $k_G$ genuine pairs and $k_I$ impostor pairs with the lowest pair quality are discarded. The remaining pairs form the test set and are used to compute the recognition error for discard rate $r$, as in the standard protocol.
While this procedure tackles a part of the \textit{Test-Set Divergence}, it still suffers from \textit{Threshold Drift} and relies on the definition of pair quality.

\subsection{Fixed-Threshold Proportional Pair Discard EDC (FT-PPD-EDC)}

While the FT-EDC addresses a problem in the error computation, PPD-EDC tackles a problem in the discarding step. 
Both approaches can be combined into the Fixed-Threshold Proportional Pair Discard EDC (FT-PPD-EDC).
FT-PPD-EDC follows the proportional pair discard procedure from PPD-EDC (Sec. \ref{sec:PPD-EDC}) and combines it with the error computation based on a fixed threshold (Sec. \ref{sec:FT-EDC}). While this integrates the strengths of both methods, it also carries forward their individual limitations.

\subsection{Rank Consistency Evaluation (RCE)}

\begin{figure}[t]
    \centering
    \includegraphics[
        width=\linewidth,
        keepaspectratio
    ]{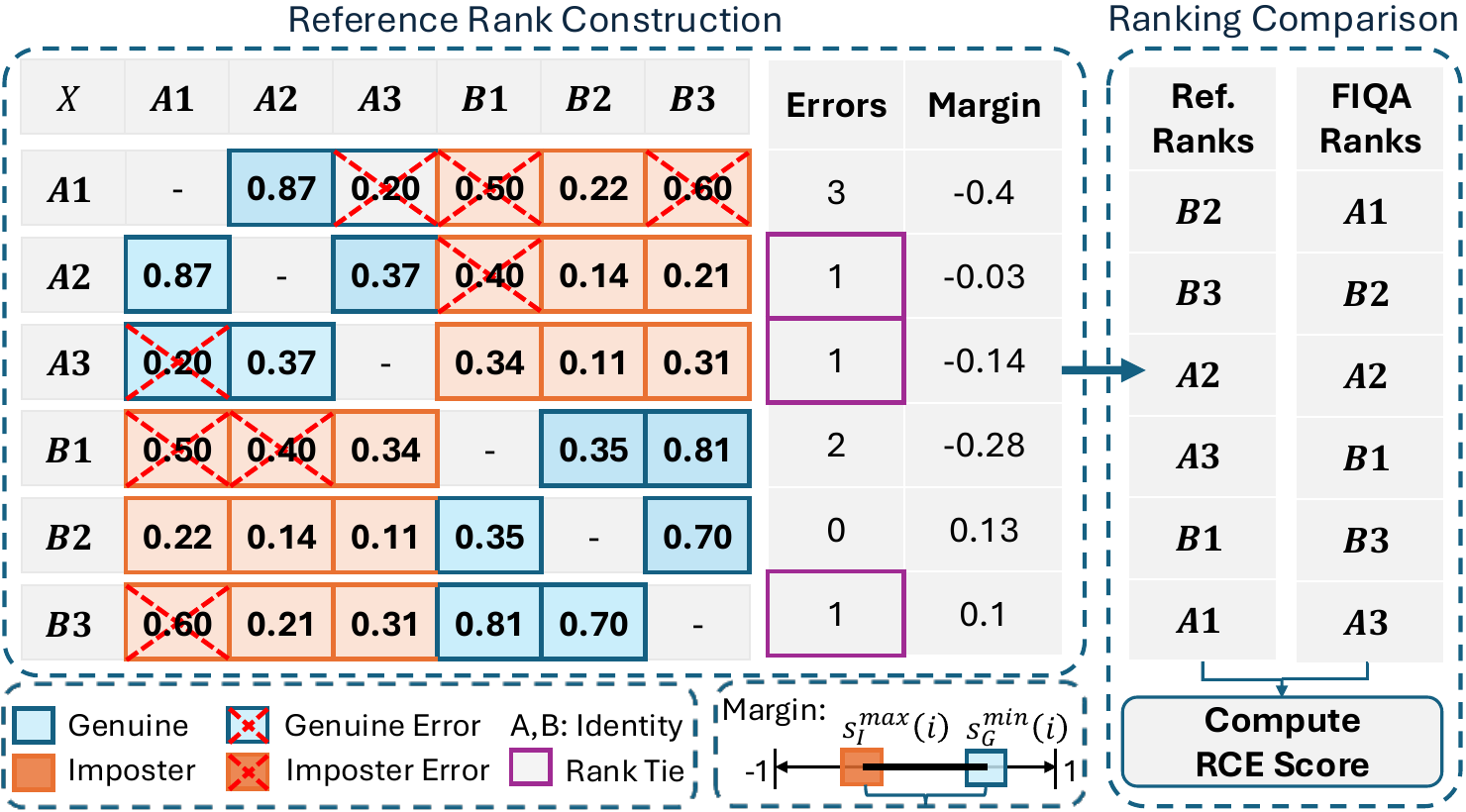}
    \caption{
    \textbf{Visualization of the RCE framework.} In Rank Consistency Evaluation (RCE), the reference ranking is constructed from recognition errors in the test set, with ties resolved based on the maximum imposter-minimum genuine margin. The final RCE score is computed as the weighted rank correlation between the reference ranking and the ranking based on the FIQA predictions.   
    }
    \label{fig:rce_framework}
    \vspace{-0.35cm}
\end{figure}

To overcome the limitations of discard-based evaluation entirely, we propose a Rank Consistency Evaluation (RCE) evaluation framework, illustrated in Figure~\ref{fig:rce_framework}. RCE first ranks the samples of the test set based on their matching errors, second, it computes a rank based on their assigned FIQA values and lastly measures the similarity between both rankings in a weighted fashion.

Let $\mathcal{X} = \{x_1, x_2, \dots, x_N\}$ denote a set of $N$ face images, with genuine ($\mathcal{G}$) and impostor ($\mathcal{I}$) pair sets. Given a similarity function $s(i,j)$ and a fixed threshold $\tau$, the error contribution {\small
\begin{equation*}
E_i = \bigl|\{(i,j)\in\mathcal{G}\mid s(i,j)<\tau\}\bigr|
      + \bigl|\{(i,j)\in\mathcal{I}\mid s(i,j)\ge\tau\}\bigr| .
\end{equation*}
} of an image $i$ is defined over the number of false non-matches and false matches it produces.

Images are ranked based on $E(i)$, where higher values indicate higher error contribution. When multiple images have identical error values in the ranking, a margin-based tie breaking is applied based on the separability of the image with its involved genuine and imposter pairs.
More precisely, for each image $i$, the minimum similarity score among genuine pairs and the maximum similarity score among impostor pairs are computed
\begin{equation}
s_G^{\min}(i) = \min_{(i,j) \in \mathcal{G}} s(i,j), \quad
s_I^{\max}(i) = \max_{(i,j) \in \mathcal{I}} s(i,j),
\end{equation}
and its margin is calculated
\begin{equation}
m(i) = s_G^{\min}(i) - s_I^{\max}(i).
\end{equation}
Among samples with equal error contribution, images with smaller margins are considered more error-prone and are therefore assigned lower ranks, as smaller margins indicate weaker separation between genuine and impostor scores and higher ambiguity in the decision boundary.

This process defines the reference ranking $R_E(i) = \mathrm{rank}(E(i))$, where higher ranks correspond to higher error contribution.
The ranking derived from recognition errors is treated as the reference for the quality, as it directly reflects the utility of an image for recognition: higher error contributions correspond to lower quality.
A FIQA model assigns a quality score $Q(i)$, which induces a predicted ranking $R_Q(i) = \mathrm{rank}(Q(i))$.

Finally, the proposed RCE metric
\begin{equation}
RCE =
1 -
\frac{
6 \sum_{i \in \mathcal{I}} w(i)\left(R_E(i) - R_Q(i)\right)^2
}{
(N^2 - 1)\sum_{i \in \mathcal{I}} w(i)
},
\end{equation}
measures agreement between the predicted ranking and the reference ranking in a weighted fashion, similar to Spearman correlation.
A high RCE value refers to a very successful FIQA algorithm that can reproduce the error-based ranking successfully, while a low RCE refers to a FIQA algorithm that assigns quality values randomly.

To account that, depending on the application, low-quality images might have a greater impact on face recognition systems, a weighting factor
\begin{equation}
w(i) =
\left(
\frac{R_E(i)}{N}
\right)^{\gamma}, \label{eq:RCEweightingfunction}
\end{equation}
is introduced that describes the importance that is assigned to each sample $i$.
Here, $\gamma$ controls the degree of weighting. A high $\gamma$ value indicates high weighting for low-quality samples, while $\gamma = 0$ leads to uniform weights and simplifies the RCE metric to the standard Spearman correlation.

While RCE eliminates \textit{Test-Set Divergence} and \textit{Threshold Drift}, as a ranking-based metric it captures only relative ordering and does not account for absolute error magnitudes; moreover, it depends on the chosen decision threshold that determines error contributions, the underlying FR model, and the weighting parameter $\gamma$.

\section{Experiments Setup}

Experiments are organized into two parts: (1) analysis of EDC limitations and (2) evaluation of EDC-based protocols and the RCE framework.

\textbf{Analysis of EDC limitations: }
To evaluate \textit{Test-Set Divergence}, experiments are reported on the Adience~\cite{6906255}, LFW~\cite{huang:inria-00321923}, and XQLFW~\cite{knoche2021cross} datasets and are independent of any FR model. FIQA methods from different recent and representative approaches are considered, including supervised regression (FaceQnet~\cite{hernandez2019faceqnet}), unsupervised estimation (SER-FIQA~\cite{Terhorst2020SERFIQUE}), recognition-based estimation (CR-FIQA~\cite{Boutros2021CRFIQAFI}), and diffusion-based estimation (eDifFIQA~\cite{10468647}). 
Pairwise comparisons of FIQA methods measure sample-set and pair-set overlap across discard rates, showing differences in the retained evaluation subsets.

To analyze \textit{Threshold Drift}, ten recent and representative FIQA methods are evaluated across four approaches: supervised regression (FaceQnet~\cite{hernandez2019faceqnet}, FaceQAN~\cite{Babnik2022FaceQANFI}), unsupervised estimation (SER-FIQA~\cite{Terhorst2020SERFIQUE}, SDD-FIQA~\cite{Ou2021SDDFIQAUF}), recognition-based methods (CR-FIQA~\cite{Boutros2021CRFIQAFI}, FROQ~\cite{Babnik2025FROQ1OF}, CLIB-FIQA~\cite{Ou_2024_CVPR}, GraFIQs~\cite{kolf2024grafiqs}), and diffusion-based approaches (eDifFIQA~\cite{10468647}, DifFIQA~\cite{babnik2023diffiqa}). The decision threshold is recomputed at each discard level to maintain $\mathrm{FMR}=10^{-3}$, following standard EDC protocol.

\textbf{Evaluation of EDC-based protocols and RCE: }
EDC-based evaluation includes four variants: EDC, FT-EDC, PPD-EDC, and FT-PPD-EDC. These are evaluated on 15 state-of-the-art FIQA methods across five widely used datasets with diverse challenges, including age variations (Adience~\cite{6906255}, CALFW~\cite{calfw}), pose variations (CPLFW~\cite{cplfw}), and unconstrained conditions (LFW~\cite{huang:inria-00321923}) and quality degradation (XQLFW~\cite{knoche2021cross}). To assess the influence of face recognition models on FIQA evaluation, experiments are conducted on four models: three CNN-based (ArcFace~\cite{Deng2018ArcFaceAA}, MagFace~\cite{Meng2021MagFaceAU}, AdaFace~\cite{Kim2022AdaFaceQA}) and one transformer-based (SwinFace~\cite{qin2023swinface}).
All CNN models use the ResNet-100 backbone, whereas SwinFace uses the Swin Transformer~\cite{qin2023swinface}. 
All models are trained on WebFace12M\footnote{\url{https://github.com/mk-minchul/AdaFace}} 
and MS1MV2\footnote{\url{https://github.com/deepinsight/insightface}}.
Recognition performance is evaluated at $\mathrm{FMR}=10^{-3}$ as per the Frontex guidelines. For comparison, pAUC is computed over the 0\%--30\% discard range, where lower values indicate better FIQA performance.


The RCE evaluation uses the same setting as the EDC-based methods for direct comparability, with $\gamma=0$ and $\gamma=3$ representing uniform and increased importance on low-quality samples, respectively.

\section{Results}


\begin{figure*}[!t]
    \centering

    \begin{subfigure}[t]{0.30\textwidth}
        \centering
        \includegraphics[width=\linewidth]{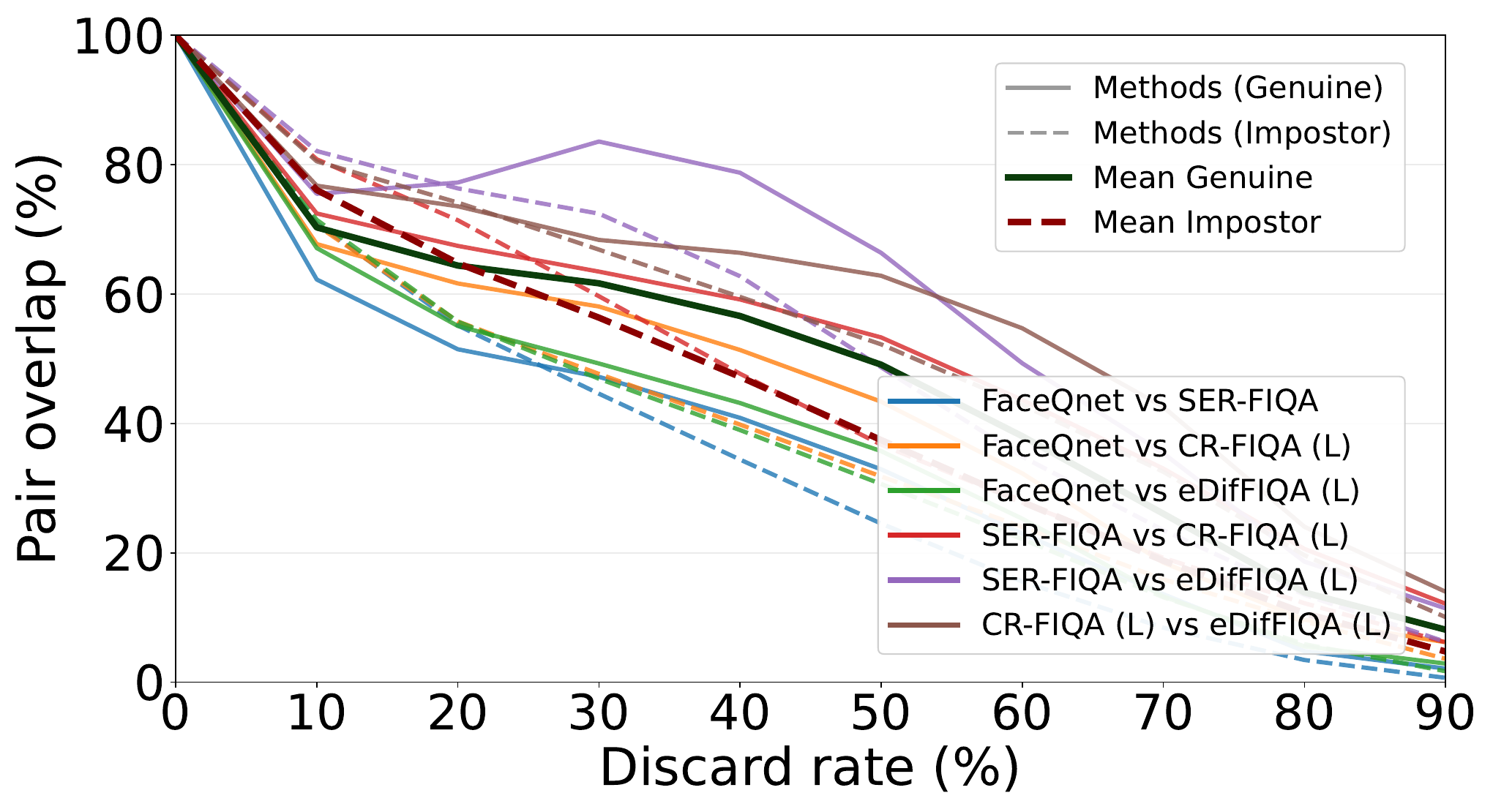}
        \vspace{-5mm}
        \caption{Adience}
        \label{fig:pair_overlap_adience}
    \end{subfigure}
    \hfill
    \begin{subfigure}[t]{0.30\textwidth}
        \centering
        \includegraphics[width=\linewidth]{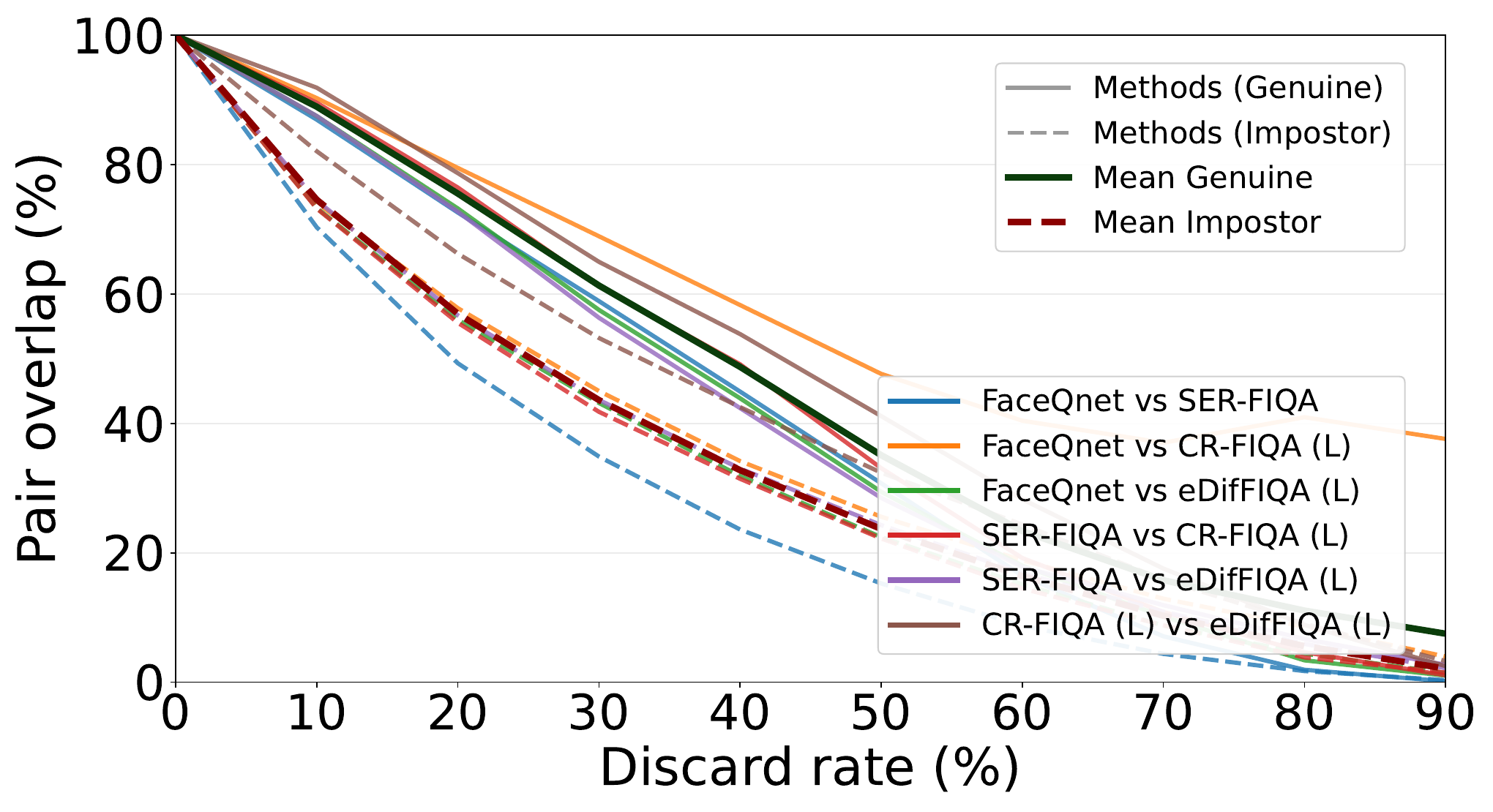}
        \vspace{-5mm}
        \caption{LFW}
        \label{fig:pair_overlap_lfw}
    \end{subfigure}
    \hfill
    \begin{subfigure}[t]{0.30\textwidth}
        \centering
        \includegraphics[width=\linewidth]{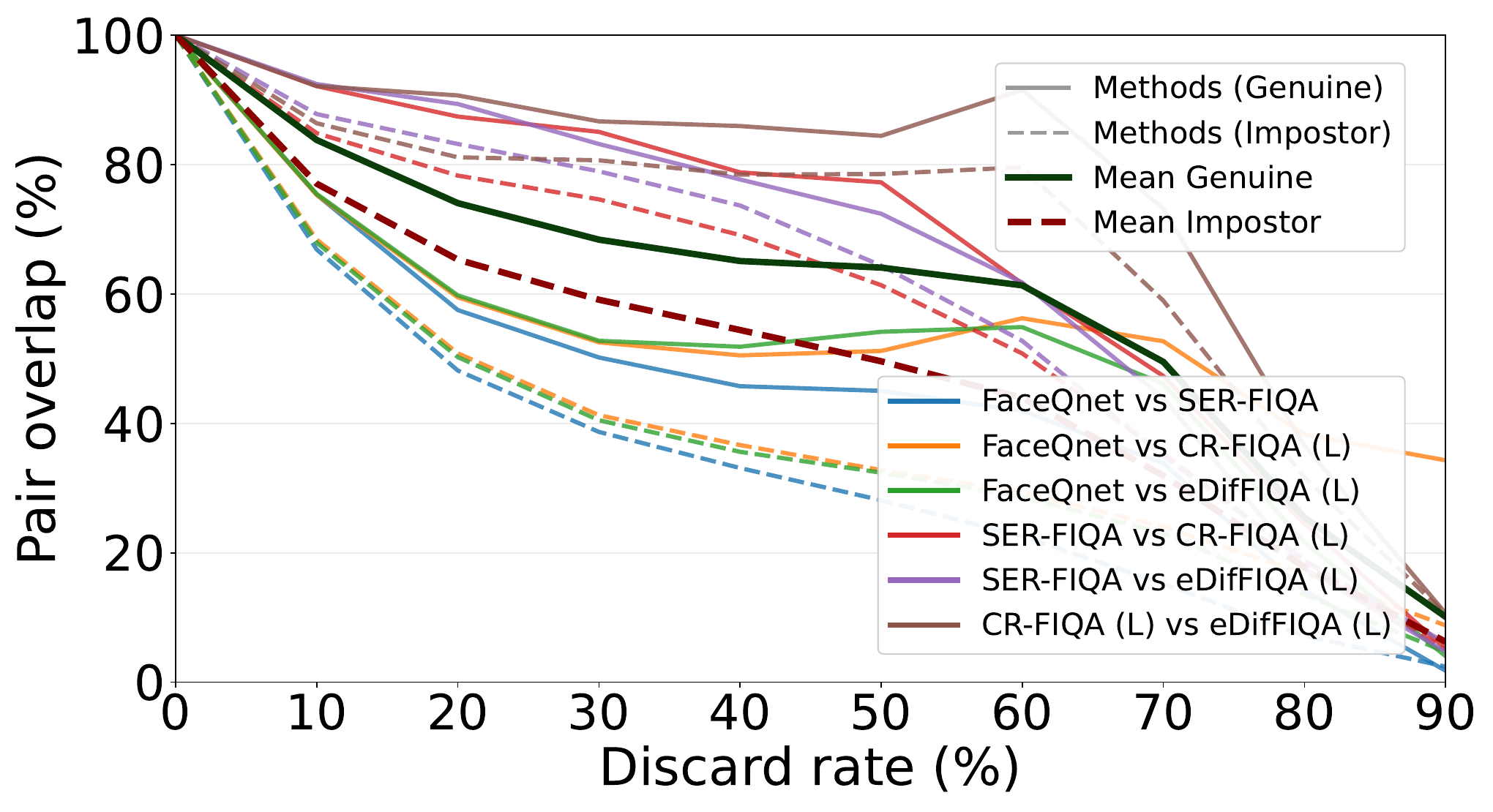}
        \vspace{-5mm}
        \caption{XQLFW}
        \label{fig:pair_overlap_xqlfw}
    \end{subfigure}

    \caption{\textbf{Retained Pair-Set Overlap under EDC.}
    The plots quantify the pairwise overlap between the test sets of sample
    pairs retained by different FIQA methods at increasing discard rates.
    The rapid decline in overlap demonstrates that, under EDC, each method
    operates on increasingly disjoint subsets of the test data.}
    \label{fig:prob_test_set_pair}
    \vspace{-0.3cm}
\end{figure*}

\subsection{Analyzing Discard-Based Evaluation Protocols}

To evaluate the EDC-based variants, \textit{Test-Set Divergence} and \textit{Threshold Drift} are analyzed to assess their effect on the consistency and comparability of FIQA evaluation.

\begin{figure*}[!t]
    \centering

    \begin{subfigure}[t]{0.30\textwidth}
        \centering
        \includegraphics[width=\linewidth]{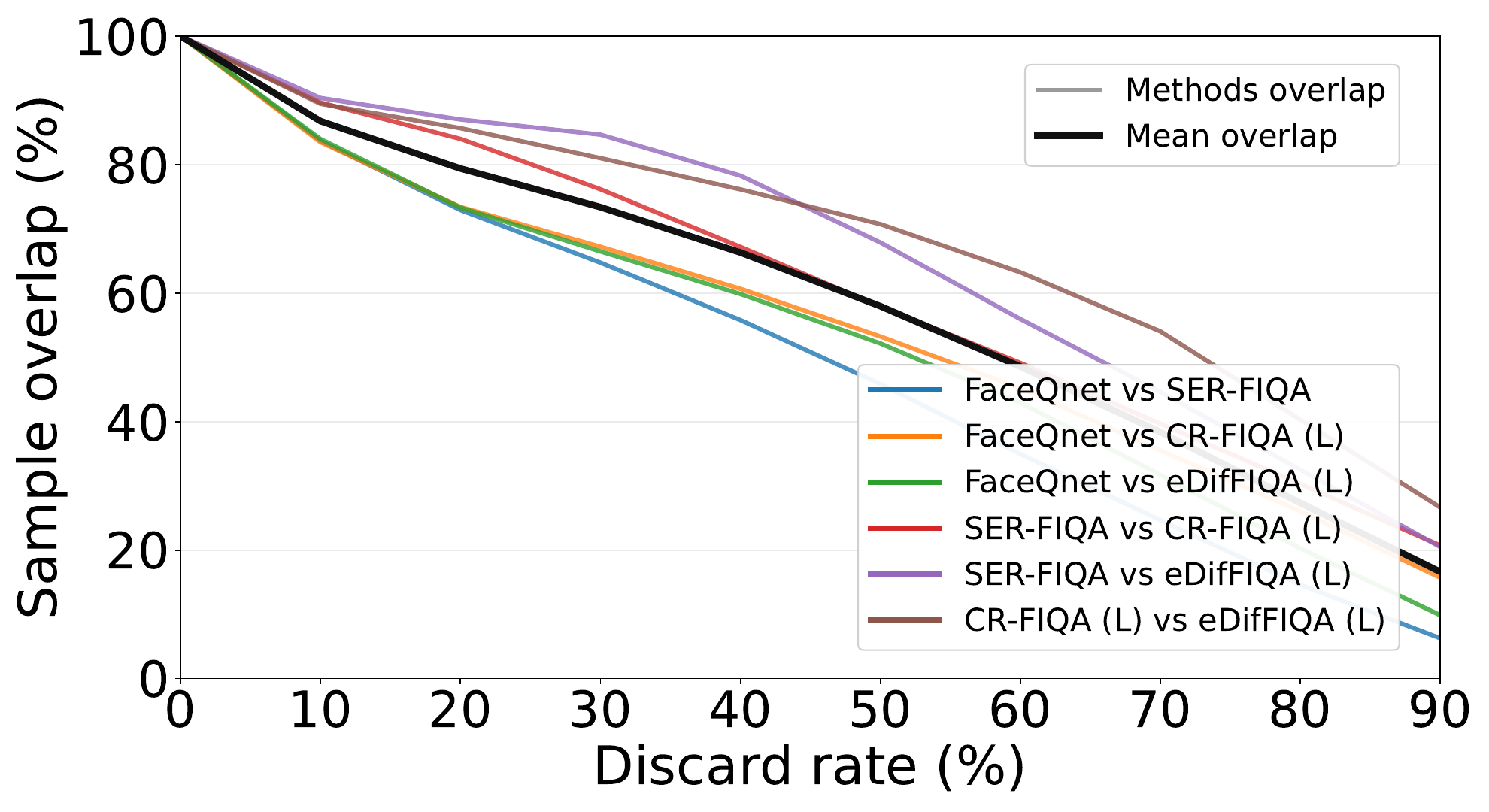}
        \vspace{-5mm}
        \caption{Adience}
        \label{fig:sample_overlap_adience}
    \end{subfigure}
    \hfill
    \begin{subfigure}[t]{0.30\textwidth}
        \centering
        \includegraphics[width=\linewidth]{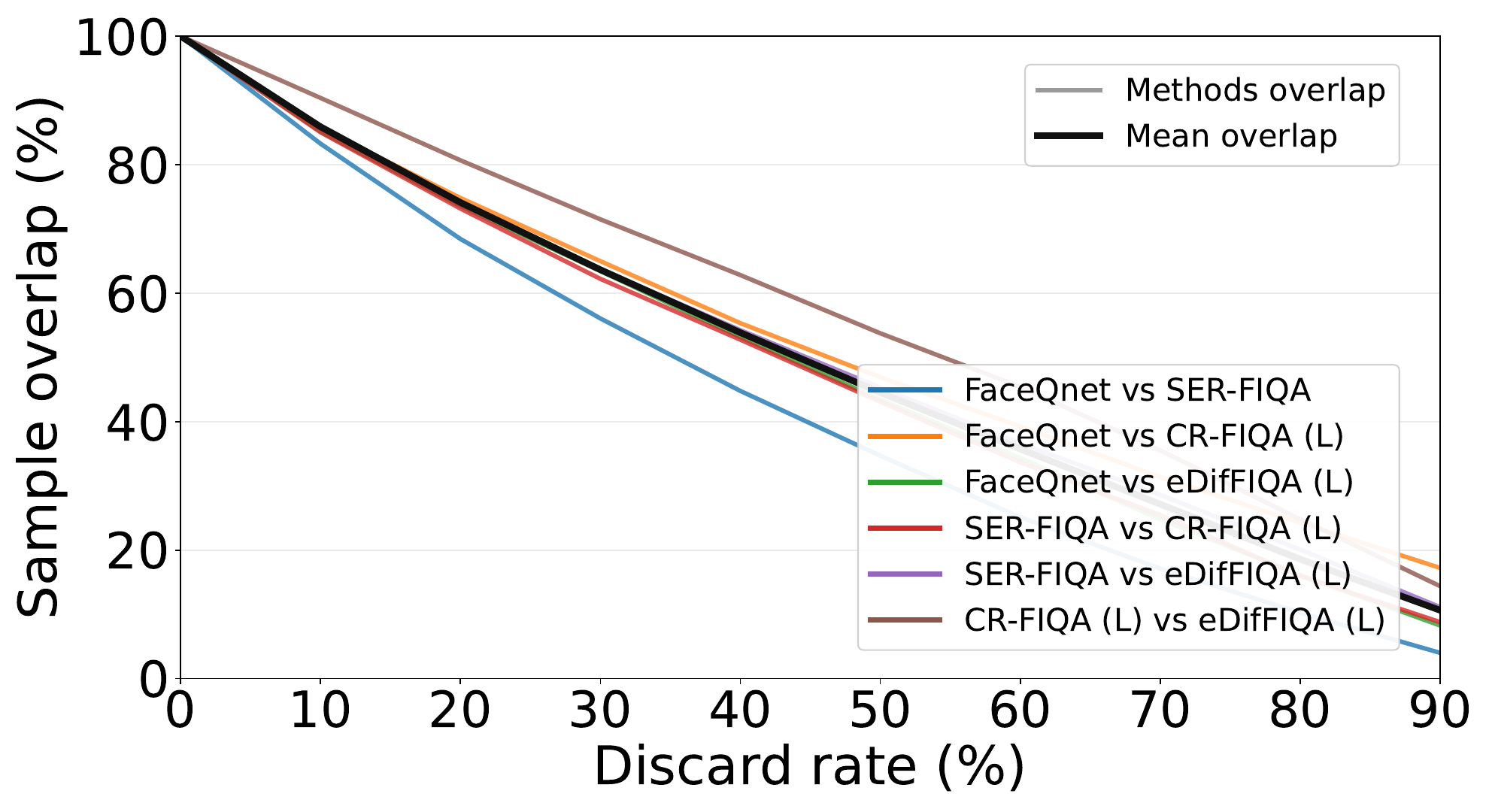}
        \vspace{-5mm}
        \caption{LFW}
        \label{fig:sample_overlap_lfw}
    \end{subfigure}
    \hfill
    \begin{subfigure}[t]{0.30\textwidth}
        \centering
        \includegraphics[width=\linewidth]{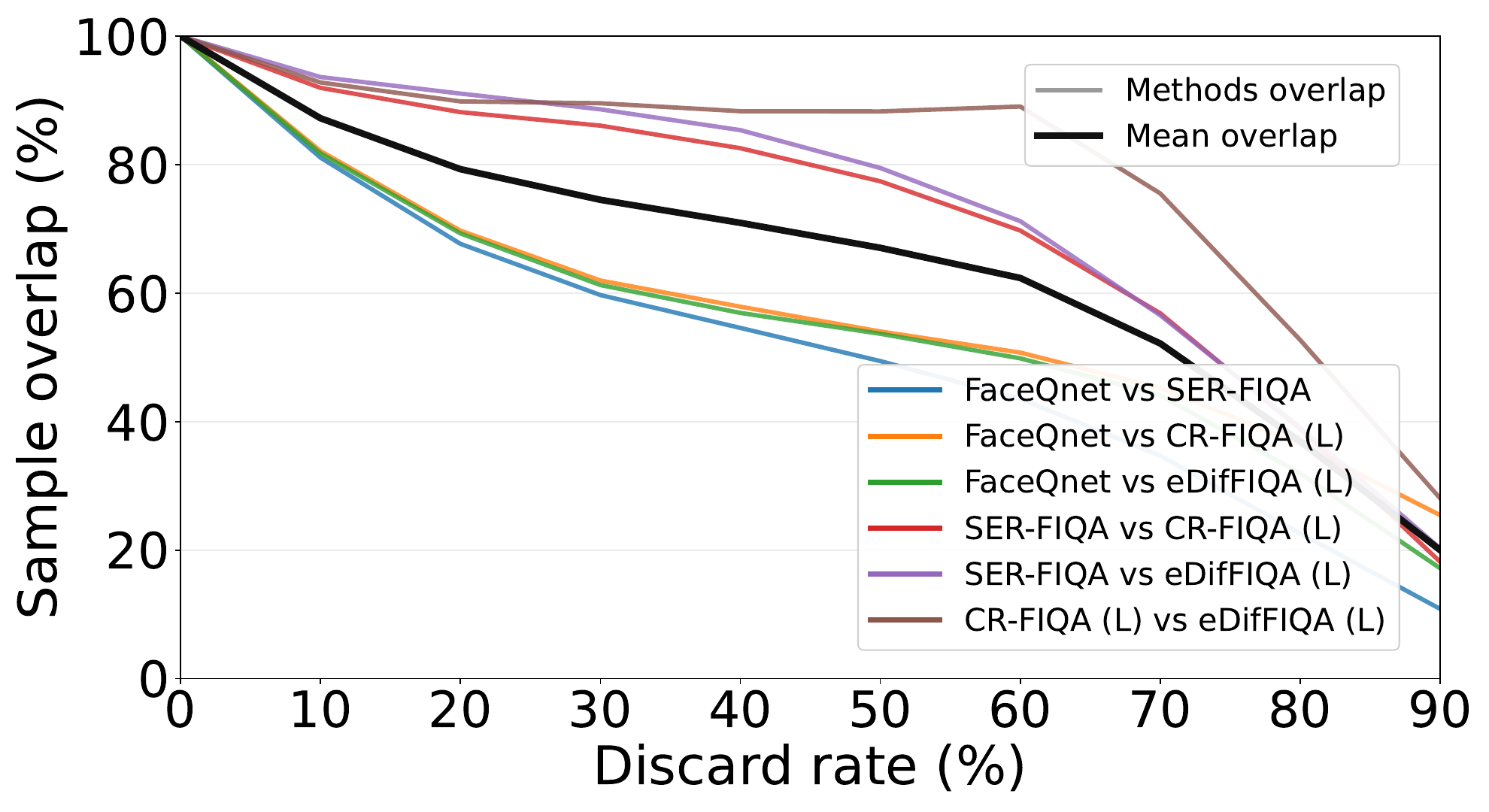}
        \vspace{-5mm}
        \caption{XQLFW}
        \label{fig:sample_overlap_xqlfw}
    \end{subfigure}

    \caption{\textbf{Retained Sample-Set Overlap under EDC.}
    Proportion of samples that remain common to the test sets of two FIQA
    methods as a function of the discard rate. As with the pair-level
    analysis, the overlap declines sharply after the initial discards,
    indicating that each FIQA method retains a markedly different subset
    of the data when evaluated using the EDC protocol.}
    \label{fig:prob_test_set_sample}
\vspace{-0.3cm}
    
\end{figure*}

\textbf{\textit{Test-Set Divergence}: }We quantify \textit{Test‑Set Divergence} under discard‑based evaluation by measuring how much the retained test data overlap when two different FIQA methods are applied at the same discard level. Figures \ref{fig:prob_test_set_pair} and \ref{fig:prob_test_set_sample} plot, as a function of the discard rate, the proportion of samples (Fig. \ref{fig:prob_test_set_sample}) and sample pairs (Fig. \ref{fig:prob_test_set_pair}) that remain in the test set for each pair of FIQA methods. The overlap monotonically declines as the discard rate increases, indicating that each FIQA algorithm preserves an increasingly distinct subset of the data at a given discard threshold. This effect is especially pronounced at the pair level, because discarding a single sample eliminates all pairs that involve that sample. The same pattern is observed for both the standard discard scheme (EDC) and the proportional‑pair discard scheme (PPD‑EDC). The corresponding overlap curves are nearly identical. Consequently, different FIQA methods operate on largely different retained samples and pairs, leading to test‑set divergence and limiting the direct comparability of their performance.

\begin{figure*}[!t]
    \centering

    \begin{subfigure}[t]{0.32\textwidth}
        \centering
        \includegraphics[width=\linewidth]{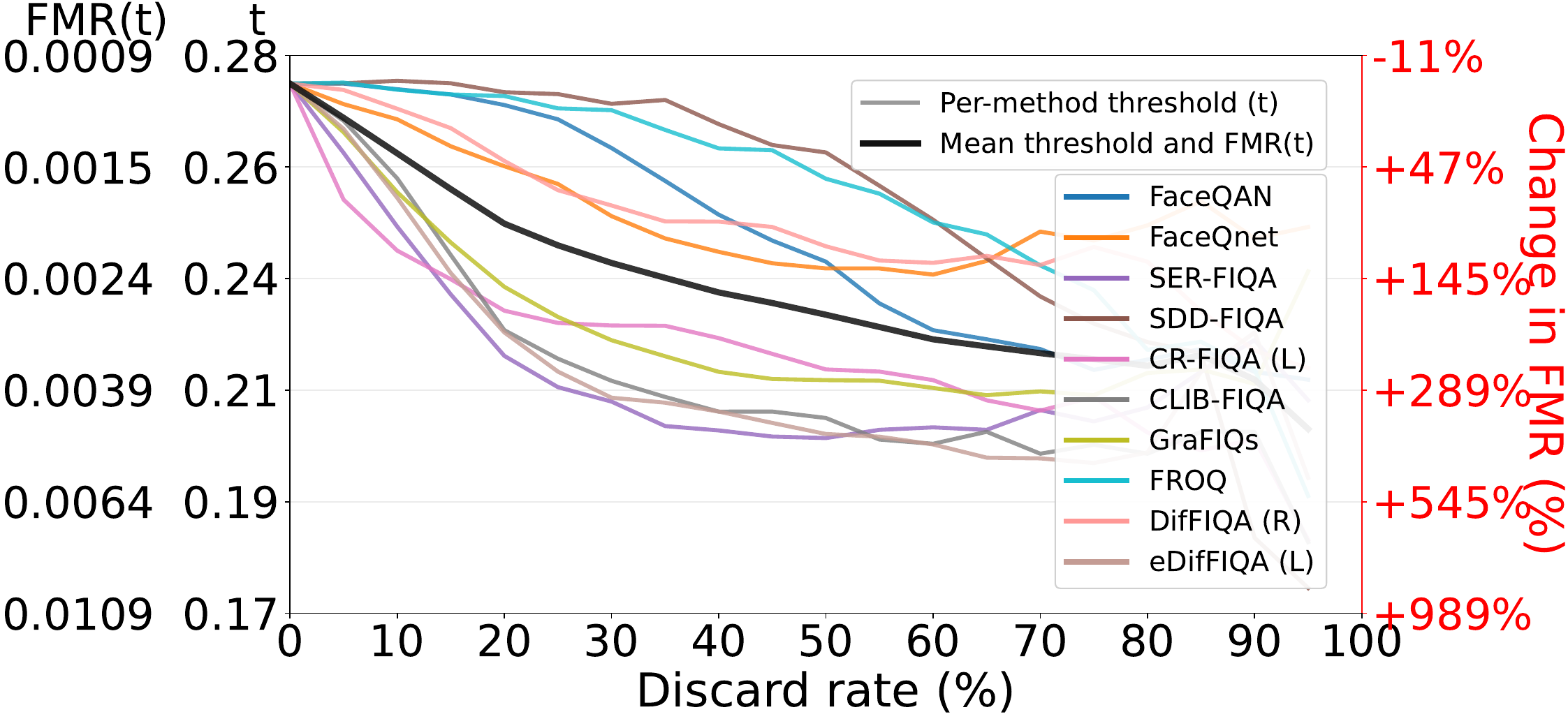}
        \vspace{-5mm}
        \caption{Adience -- AdaFace}
        \label{fig:thr_adaface_adience}
    \end{subfigure}
    \hfill
    \begin{subfigure}[t]{0.32\textwidth}
        \centering
        \includegraphics[width=\linewidth]{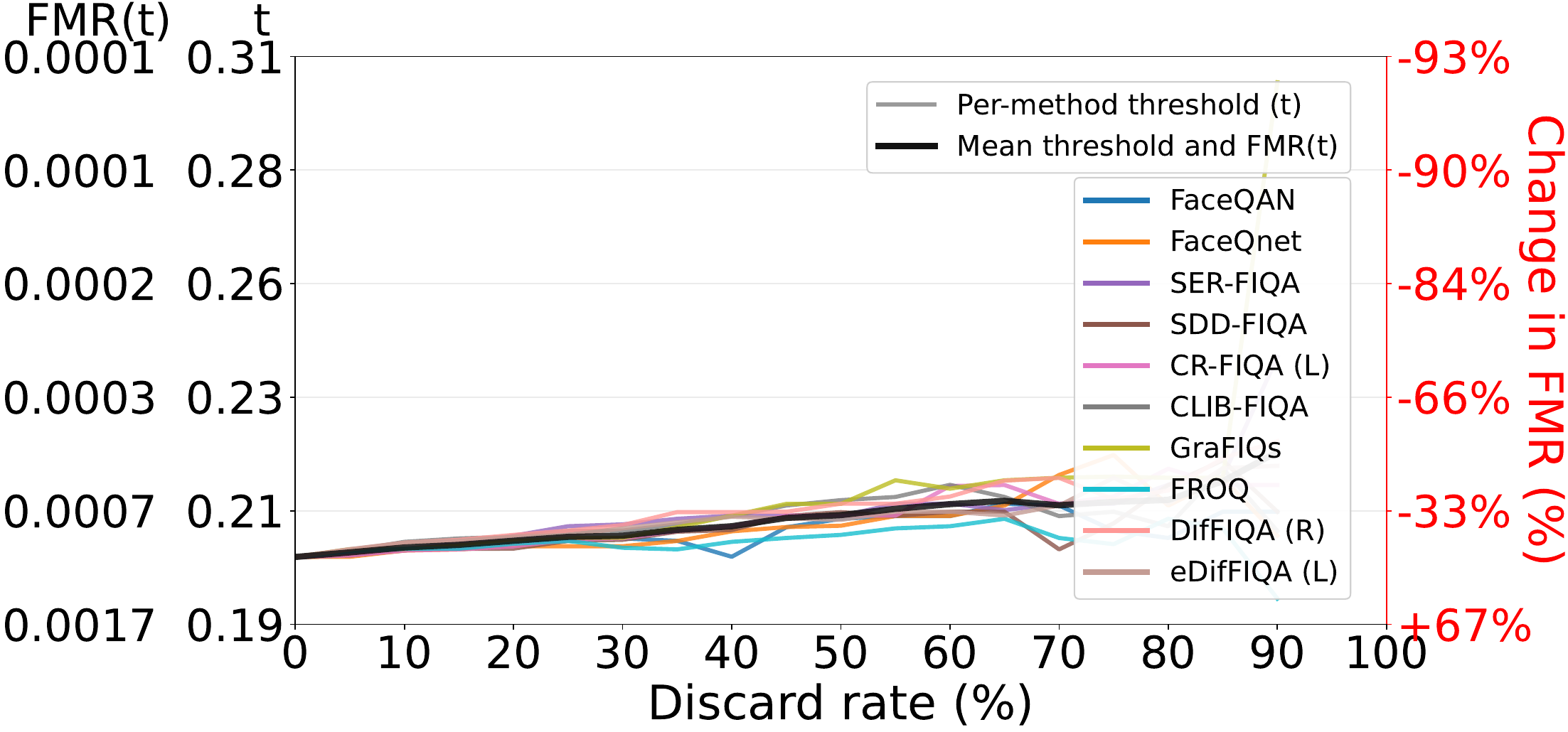}
        \vspace{-5mm}
        \caption{LFW -- AdaFace}
        \label{fig:thr_adaface_lfw}
    \end{subfigure}
    \hfill
    \begin{subfigure}[t]{0.32\textwidth}
        \centering
        \includegraphics[width=\linewidth]{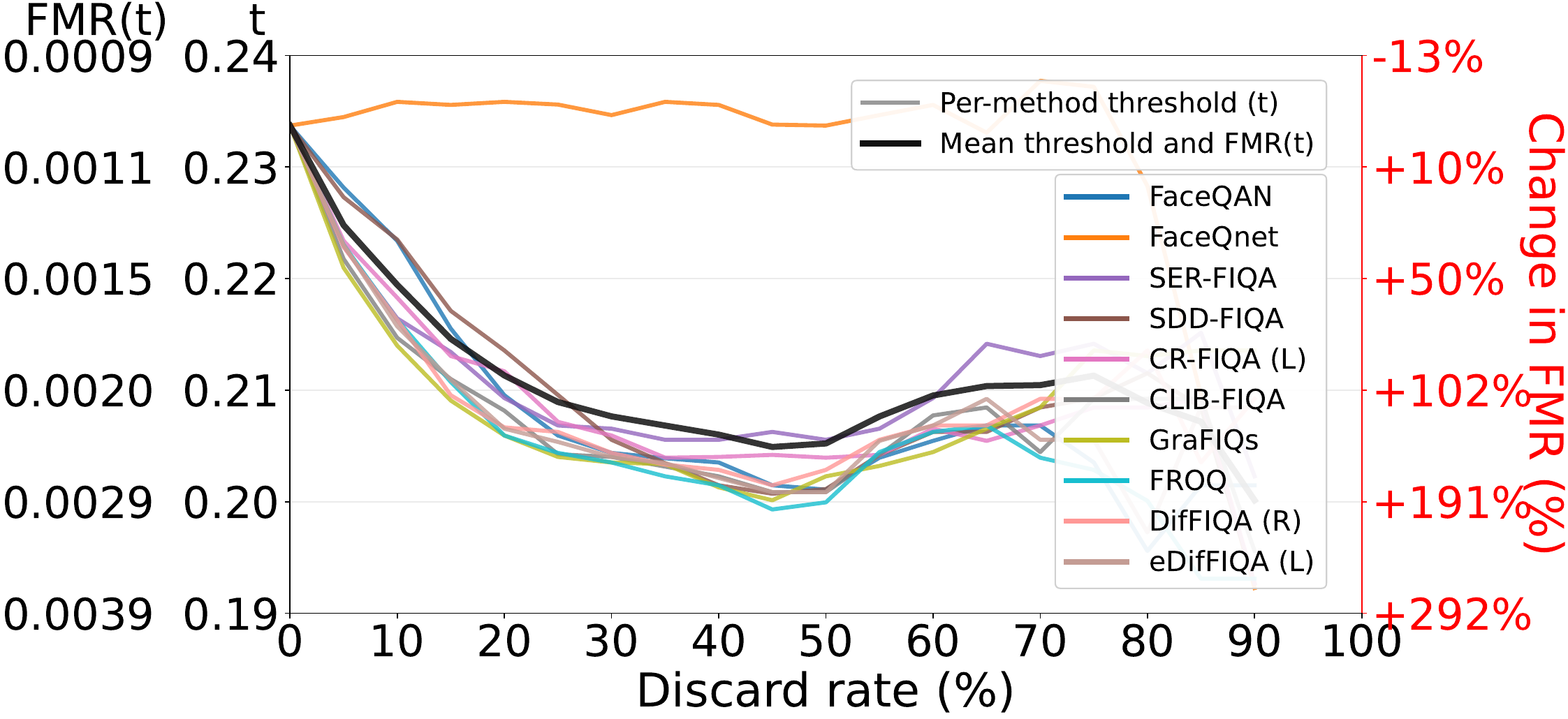}
        \vspace{-5mm}
        \caption{XQLFW -- AdaFace}
        \label{fig:thr_adaface_xqlfw}
    \end{subfigure}

    \vspace{-0.1cm}

    \begin{subfigure}[t]{0.32\textwidth}
        \centering
        \includegraphics[width=\linewidth]{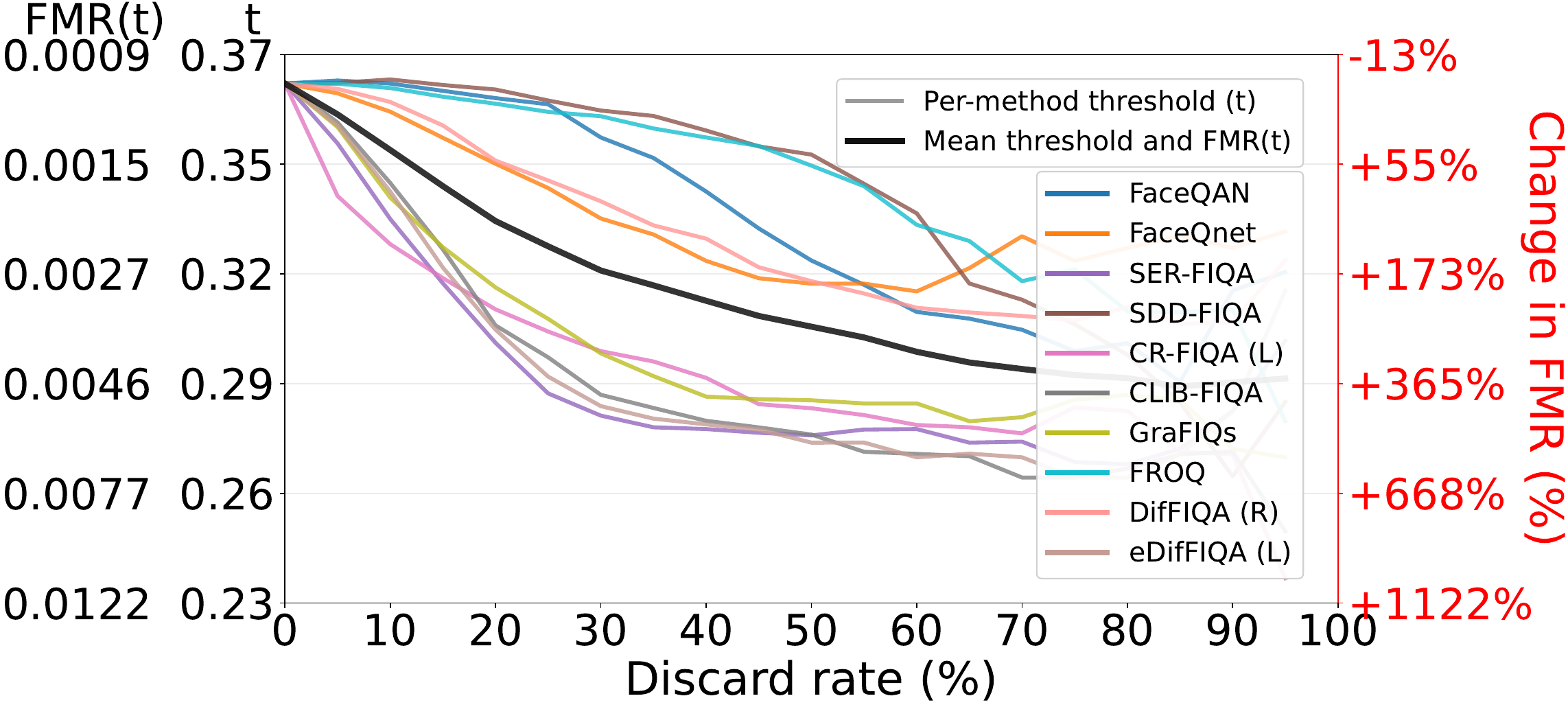}
        \vspace{-5mm}
        \caption{Adience -- ArcFace}
        \label{fig:thr_arcface_adience}
    \end{subfigure}
    \hfill
    \begin{subfigure}[t]{0.32\textwidth}
        \centering
        \includegraphics[width=\linewidth]{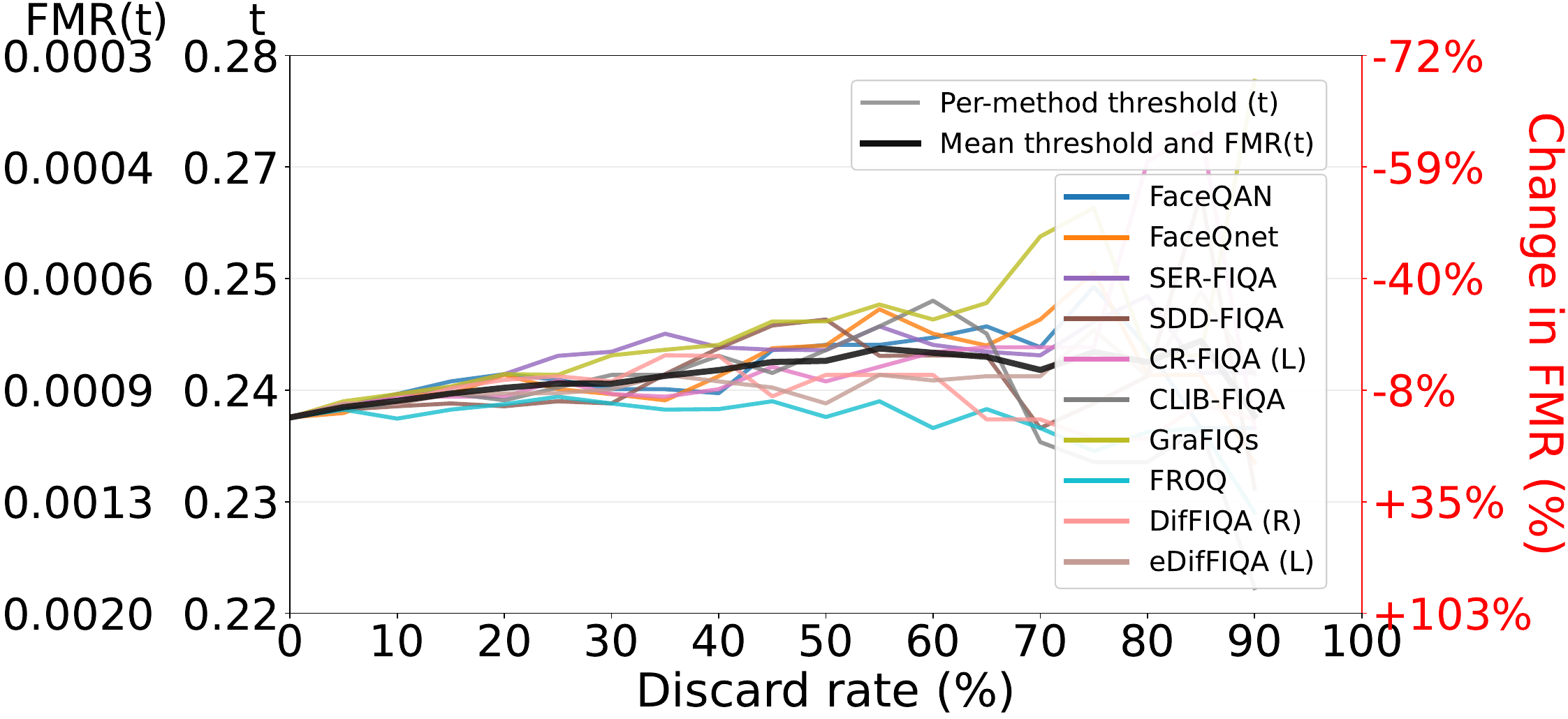}
        \vspace{-5mm}
        \caption{LFW -- ArcFace}
        \label{fig:thr_arcface_lfw}
    \end{subfigure}
    \hfill
    \begin{subfigure}[t]{0.32\textwidth}
        \centering
        \includegraphics[width=\linewidth]{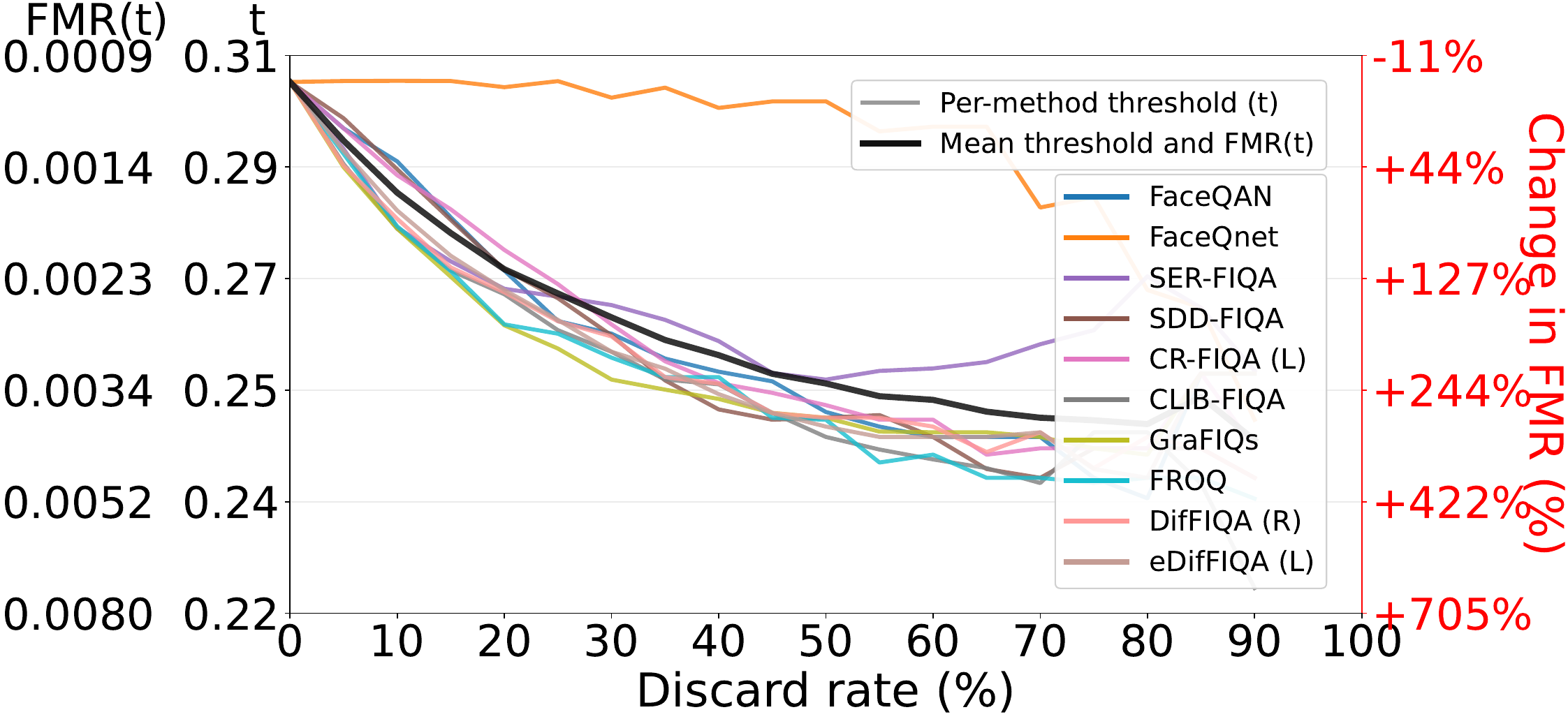}
        \vspace{-5mm}
        \caption{XQLFW -- ArcFace}
        \label{fig:thr_arcface_xqlfw}
    \end{subfigure}

    \caption{\textbf{Threshold Drift under EDC.}
    The plots show how the decision threshold evolves with increasing
    discard rate across datasets and face recognition models.
    To quantify its impact, the threshold shifts are translated into
    corresponding changes in FMR on the full dataset. The results show that
    EDC drives different methods to operate at substantially different
    decision points, undermining direct comparability.}
    \label{fig:prob_thr_grid}
    \vspace{-0.35cm}
\end{figure*}

\textbf{\textit{Threshold Drift}: }
In the EDC evaluation protocol, the decision threshold is recomputed after each discard step to ensure the retained subset meets the target FMR. Figure \ref{fig:prob_thr_grid} shows how this threshold varies with the discard rate for various FIQA methods, datasets, and FR models. The threshold changes continuously as more samples are discarded: lowering the threshold raises the FMR, while raising it reduces the FMR. Consequently, if the threshold obtained at a given discard level is applied to the full (un‑discarded) test set (shown in Figure \ref{fig:prob_thr_grid} in red), the resulting FMR can deviate dramatically from the intended operating point—often by several hundred percent. This behavior is observed for both the standard‑discard scheme (EDC) and the proportional‑pair discard scheme (PPD‑EDC), indicating that the operating point shifts with the discard level. As a result, different FIQA methods end up operating at substantially different operation points, which significantly limits the direct comparability.

\textbf{EDC-based Evaluation: }To evaluate the proposed EDC-based variants with the standard protocol, Table~\ref{tab:fiqa_pauc} reports the performance of FIQA methods across five datasets under four EDC-based protocols. The results show that the relative ranking of FIQA methods is highly consistent across all protocol variants. FROQ achieves the best average rank across all protocols, followed by the CLIB-FIQA and eDifFIQA (L) and (M) or ViT-FIQA, while methods such as FaceQnet and FaceQGen consistently rank lower.

Across EDC, FT-EDC, PPD-EDC, and FT-PPD-EDC, the relative ranking of FIQA methods remains largely consistent. Although the numerical values of FNMR and CER differ slightly across protocols, these variations do not result in significant changes in ordering. 

This stability, however, does not mitigate fundamental limitations inherent to discard-based evaluation. All EDC variants, including the adapted protocols, suffer from \textit{Test-Set Divergence} and \textit{Threshold Drift}. Consequently, despite consistent ordinal performance, the absolute reliability of FIQA comparisons remains compromised, as methods operate on non-identical test subsets under non-equivalent decision thresholds.


\subsection{Analyzing Rank-Based Evaluation Protocol}

Since even refined EDC variants fail to resolve the fundamental limitations of discard-based evaluation, such as \textit{Test-Set Divergence} and \textit{Threshold Drift}, we propose Rank Consistency Evaluation (RCE) as a fundamentally different, ranking-based evaluation framework for FIQA.  

To assess state-of-the-art FIQA methods under stable, comparable conditions, Table~\ref{tab:fiqa_ranking} reports RCE scores under two weighting schemes: $\gamma = 0$ (equal sample importance) and $\gamma = 3$ (higher weight on low‑quality samples). Crucially, RCE evaluates all methods on the full test set using a single, fixed decision threshold computed at the target FMR. This eliminates sample discarding and threshold drift, ensuring that all FIQA methods are assessed under identical conditions; in stark contrast to EDC-based protocols, which vary test sets and operating points with each discard step.

As a result, the observed ranking of FIQA methods is protocol-dependent. Compared to Table~\ref{tab:fiqa_pauc} (EDC-based), RCE yields markedly different orderings: methods that lead under EDC, such as FROQ, drop to mid-tier performance across all FR models under RCE. Conversely, eDifFIQA (L) emerges as the most consistent and top-performing method under RCE, achieving the best average rank across both $\gamma$ settings. CLIB‑FIQA remains strong under both protocols, while FaceQnet consistently ranks low. Notably, recent 2025 methods, including FROQ~\cite{Babnik2025FROQ1OF} and ViT‑FIQA~\cite{atzori2025vit}, which excel under EDC, fall to mid- or lower-rank positions under RCE. This demonstrates that performance gains under discard-based evaluation do not necessarily reflect robust, generalizable quality assessment.

In summary, RCE provides a more stable, comparable evaluation environment by design; avoiding the variability inherent in EDC, and thus, offers a more reliable basis for ranking FIQA methods.

\begin{sidewaystable*}[p]
\centering
\scriptsize
\setlength{\tabcolsep}{3pt}
\renewcommand{\arraystretch}{1.08}
\resizebox{\textheight}{!}{%
\begin{tabular}{ll|r|r|r|r|r|r|r|r|r|r|r|r|r|r|r|r|r|r|r|r|r|r|r|r}
\toprule
\multirow{3}{*}{FR} & \multirow{3}{*}{FIQA} & \multicolumn{4}{c|}{Adience~\cite{6906255}}  & \multicolumn{4}{c|}{CALFW~\cite{calfw}} & \multicolumn{4}{c|}{CPLFW~\cite{cplfw}} & \multicolumn{4}{c|}{LFW~\cite{huang:inria-00321923}} & \multicolumn{4}{c|}{XQLFW~\cite{knoche2021cross}}  & \multicolumn{4}{c}{Summary} \\
 &  & \multicolumn{1}{c|}{EDC} & \multicolumn{1}{c|}{FT} & \multicolumn{1}{c|}{PPD} & \multicolumn{1}{c|}{FT-PPD} & \multicolumn{1}{c|}{EDC} & \multicolumn{1}{c|}{FT} & \multicolumn{1}{c|}{PPD} & \multicolumn{1}{c|}{FT-PPD} & \multicolumn{1}{c|}{EDC} & \multicolumn{1}{c|}{FT} & \multicolumn{1}{c|}{PPD} & \multicolumn{1}{c|}{FT-PPD} & \multicolumn{1}{c|}{EDC} & \multicolumn{1}{c|}{FT} & \multicolumn{1}{c|}{PPD} & \multicolumn{1}{c|}{FT-PPD} & \multicolumn{1}{c|}{EDC} & \multicolumn{1}{c|}{FT} & \multicolumn{1}{c|}{PPD} & \multicolumn{1}{c|}{FT-PPD} & \multicolumn{1}{c|}{EDC} & \multicolumn{1}{c|}{FT-EDC} & \multicolumn{1}{c|}{PPD-EDC} & \multicolumn{1}{c}{FT-PPD-EDC} \\
 &  & FNMR & CER & FNMR & CER & FNMR & CER & FNMR & CER & FNMR & CER & FNMR & CER & FNMR & CER & FNMR & CER & FNMR & CER & FNMR & CER & Avg.\ Rank & Avg.\ Rank & Avg.\ Rank & Avg.\ Rank \\
\midrule
\multirow{18}{*}{\rotatebox{90}{AdaFace~\cite{Kim2022AdaFaceQA}}} & FaceQnet~\cite{hernandez2019faceqnet} & (19) 0.0138 & (18) 0.0073 & (19) 0.0139 & (19) 0.0072 & (17) 0.0131 & (17) 0.0067 & (17) 0.0135 & (17) 0.0069 & (18) 0.0263 & (18) 0.0133 & (18) 0.0270 & (18) 0.0137 & (19) 0.0013 & (19) 0.0008 & (19) 0.0013 & (19) 0.0008 & (19) 0.0488 & (19) 0.0243 & (19) 0.0497 & (19) 0.0248 & (16) 18.40 & (17) 18.20 & (15) 18.40 & (16) 18.40 \\
 & SER-FIQA~\cite{Terhorst2020SERFIQUE} & (6) 0.0104 & (11) 0.0057 & (6) 0.0112 & (12) 0.0062 & (16) 0.0126 & (16) 0.0065 & (14) 0.0128 & (14) 0.0066 & (15) 0.0185 & (15) 0.0094 & (11) 0.0195 & (11) 0.0099 & (17) 0.0007 & (17) 0.0005 & (17) 0.0006 & (17) 0.0004 & (13) 0.0218 & (15) 0.0131 & (8) 0.0233 & (9) 0.0133 & (13) 13.40 & (15) 14.80 & (10) 11.20 & (13) 12.60 \\
 & PFE~\cite{shi2019probabilistic} & (11) 0.0107 & (7) 0.0055 & (10) 0.0114 & (7) 0.0059 & (12) 0.0120 & (12) 0.0061 & (7) 0.0125 & (7) 0.0064 & (16) 0.0186 & (16) 0.0095 & (17) 0.0213 & (17) 0.0108 & (7) 0.0005 & (7) 0.0004 & (7) 0.0004 & (6) 0.0004 & (6) 0.0197 & (5) 0.0111 & (15) 0.0287 & (14) 0.0158 & (10) 10.40 & (9) 9.40 & (10) 11.20 & (11) 10.20 \\
 & MagFace~\cite{Meng2021MagFaceAU} & (12) 0.0108 & (8) 0.0057 & (12) 0.0116 & (8) 0.0061 & (5) 0.0116 & (5) 0.0060 & (5) 0.0123 & (5) 0.0063 & (14) 0.0183 & (14) 0.0093 & (16) 0.0207 & (16) 0.0105 & (5) 0.0004 & (4) 0.0004 & (8) 0.0004 & (7) 0.0004 & (18) 0.0393 & (18) 0.0209 & (18) 0.0409 & (18) 0.0213 & (11) 10.80 & (11) 9.80 & (11) 11.80 & (12) 10.80 \\
 & SDD-FIQA~\cite{Ou2021SDDFIQAUF} & (8) 0.0105 & (6) 0.0054 & (8) 0.0113 & (5) 0.0058 & (8) 0.0118 & (8) 0.0060 & (12) 0.0126 & (11) 0.0065 & (12) 0.0178 & (12) 0.0090 & (13) 0.0204 & (13) 0.0104 & (2) \underline{0.0004} & (2) \underline{0.0004} & (2) \underline{0.0004} & (2) \underline{0.0003} & (16) 0.0279 & (16) 0.0158 & (16) 0.0320 & (16) 0.0172 & (8) 9.20 & (7) 8.80 & (9) 10.20 & (9) 9.40 \\
 & LightQNet~\cite{9528058} & (5) 0.0103 & (3) 0.0054 & (4) 0.0111 & (4) 0.0057 & (11) 0.0119 & (11) 0.0061 & (11) 0.0126 & (12) 0.0065 & (13) 0.0182 & (13) 0.0093 & (15) 0.0206 & (15) 0.0105 & (16) 0.0006 & (15) 0.0005 & (6) 0.0004 & (8) 0.0004 & (11) 0.0211 & (11) 0.0125 & (10) 0.0235 & (8) 0.0132 & (12) 11.20 & (12) 10.60 & (6) 9.20 & (9) 9.40 \\
 & FaceQGen~\cite{HernandezOrtega2021FaceQgenSD} & (17) 0.0122 & (17) 0.0064 & (17) 0.0126 & (16) 0.0065 & (18) 0.0135 & (18) 0.0069 & (18) 0.0136 & (18) 0.0070 & (17) 0.0189 & (17) 0.0096 & (14) 0.0205 & (14) 0.0104 & (18) 0.0012 & (18) 0.0007 & (18) 0.0010 & (18) 0.0007 & (15) 0.0222 & (13) 0.0131 & (14) 0.0286 & (15) 0.0161 & (15) 17.00 & (16) 16.60 & (14) 16.20 & (15) 16.20 \\
 & FaceQAN~\cite{Babnik2022FaceQANFI} & (16) 0.0116 & (13) 0.0060 & (14) 0.0121 & (13) 0.0062 & (14) 0.0124 & (14) 0.0063 & (16) 0.0131 & (16) 0.0067 & (11) 0.0175 & (11) 0.0089 & (12) 0.0199 & (12) 0.0101 & (10) 0.0005 & (11) 0.0004 & (11) 0.0005 & (10) 0.0004 & (17) 0.0303 & (17) 0.0173 & (17) 0.0326 & (17) 0.0175 & (14) 13.60 & (14) 13.20 & (13) 14.00 & (14) 13.60 \\
 & CR-FIQA (L)~\cite{Boutros2021CRFIQAFI} & (13) 0.0111 & (14) 0.0060 & (16) 0.0122 & (17) 0.0068 & (4) 0.0116 & (4) 0.0059 & (4) 0.0122 & (4) 0.0063 & (8) 0.0168 & (8) 0.0085 & (8) 0.0183 & (8) 0.0093 & (11) 0.0005 & (10) 0.0004 & (9) 0.0004 & (9) 0.0004 & (10) 0.0211 & (10) 0.0124 & (7) 0.0230 & (7) 0.0130 & (8) 9.20 & (8) 9.20 & (5) 8.80 & (7) 9.00 \\
 & DifFIQA~\cite{babnik2023diffiqa} & (15) 0.0113 & (16) 0.0061 & (13) 0.0120 & (14) 0.0064 & (9) 0.0118 & (9) 0.0061 & (8) 0.0125 & (8) 0.0064 & (7) 0.0167 & (7) 0.0085 & (7) 0.0181 & (7) 0.0092 & (8) 0.0005 & (8) 0.0004 & (13) 0.0005 & (12) 0.0004 & (8) 0.0198 & (8) 0.0118 & (6) 0.0224 & (5) 0.0127 & (9) 9.40 & (10) 9.60 & (7) 9.40 & (8) 9.20 \\
 & DifFIQA (R)~\cite{babnik2023diffiqa} & (4) 0.0103 & (4) 0.0054 & (7) 0.0112 & (6) 0.0058 & (10) 0.0118 & (10) 0.0061 & (6) 0.0124 & (6) 0.0063 & (4) 0.0159 & (3) 0.0081 & (5) 0.0175 & (5) 0.0089 & (15) 0.0006 & (16) 0.0005 & (15) 0.0005 & (15) 0.0004 & (4) 0.0185 & (4) 0.0110 & (1) \textbf{0.0208} & (1) \textbf{0.0119} & (6) 7.40 & (5) 7.40 & (3) 6.80 & (5) 6.60 \\
 & CLIB-FIQA~\cite{Ou_2024_CVPR} & (10) 0.0107 & (12) 0.0057 & (11) 0.0115 & (11) 0.0061 & (3) 0.0113 & (3) 0.0058 & (3) 0.0122 & (3) 0.0062 & (3) 0.0159 & (4) 0.0081 & (4) 0.0174 & (4) 0.0089 & (4) 0.0004 & (6) 0.0004 & (3) 0.0004 & (4) 0.0004 & (3) 0.0175 & (3) 0.0104 & (2) \underline{0.0210} & (2) \underline{0.0121} & (2) \underline{4.60} & (2) \underline{5.60} & (2) \underline{4.60} & (2) \underline{4.80} \\
 & GraFIQs~\cite{kolf2024grafiqs} & (14) 0.0112 & (15) 0.0060 & (15) 0.0121 & (15) 0.0065 & (15) 0.0125 & (15) 0.0064 & (15) 0.0129 & (15) 0.0066 & (10) 0.0172 & (10) 0.0088 & (10) 0.0191 & (10) 0.0097 & (14) 0.0006 & (14) 0.0004 & (16) 0.0006 & (16) 0.0004 & (1) \textbf{0.0165} & (1) \textbf{0.0096} & (12) 0.0251 & (12) 0.0143 & (11) 10.80 & (13) 11.00 & (12) 13.60 & (14) 13.60 \\
 & eDifFIQA (S)~\cite{10468647} & (9) 0.0106 & (9) 0.0057 & (9) 0.0113 & (9) 0.0061 & (6) 0.0117 & (7) 0.0060 & (10) 0.0126 & (10) 0.0064 & (6) 0.0164 & (6) 0.0083 & (6) 0.0180 & (6) 0.0091 & (9) 0.0005 & (9) 0.0004 & (14) 0.0005 & (13) 0.0004 & (9) 0.0209 & (9) 0.0124 & (9) 0.0235 & (10) 0.0133 & (7) 7.80 & (6) 8.00 & (8) 9.60 & (10) 9.60 \\
 & eDifFIQA (M)~\cite{10468647} & (7) 0.0104 & (10) 0.0057 & (5) 0.0112 & (10) 0.0061 & (1) \textbf{0.0105} & (1) \textbf{0.0054} & (1) \textbf{0.0114} & (1) \textbf{0.0059} & (2) \underline{0.0157} & (2) \underline{0.0080} & (2) \underline{0.0173} & (2) \underline{0.0088} & (12) 0.0005 & (12) 0.0004 & (10) 0.0005 & (11) 0.0004 & (12) 0.0215 & (12) 0.0127 & (3) 0.0217 & (3) 0.0123 & (5) 6.80 & (5) 7.40 & (1) \textbf{4.20} & (4) 5.40 \\
 & eDifFIQA (L)~\cite{10468647} & (3) 0.0101 & (5) 0.0054 & (2) \underline{0.0106} & (3) 0.0057 & (2) \underline{0.0112} & (2) \underline{0.0057} & (2) \underline{0.0121} & (2) \underline{0.0062} & (5) 0.0159 & (5) 0.0081 & (3) 0.0174 & (3) 0.0088 & (13) 0.0005 & (13) 0.0004 & (12) 0.0005 & (14) 0.0004 & (5) 0.0189 & (6) 0.0112 & (4) 0.0221 & (4) 0.0125 & (3) 5.60 & (4) 6.20 & (2) \underline{4.60} & (3) 5.20 \\
 & FROQ~\cite{Babnik2025FROQ1OF} & (1) \textbf{0.0092} & (1) \textbf{0.0048} & (1) \textbf{0.0102} & (1) \textbf{0.0053} & (13) 0.0122 & (13) 0.0062 & (13) 0.0127 & (13) 0.0065 & (1) \textbf{0.0155} & (1) \textbf{0.0079} & (1) \textbf{0.0171} & (1) \textbf{0.0087} & (1) \textbf{0.0003} & (1) \textbf{0.0003} & (1) \textbf{0.0003} & (1) \textbf{0.0003} & (2) \underline{0.0168} & (2) \underline{0.0100} & (5) 0.0222 & (6) 0.0127 & (1) \textbf{3.60} & (1) \textbf{3.60} & (1) \textbf{4.20} & (1) \textbf{4.40} \\
 & ViT-FIQA~\cite{atzori2025vit} & (2) \underline{0.0101} & (2) \underline{0.0053} & (3) 0.0110 & (2) \underline{0.0057} & (7) 0.0117 & (6) 0.0060 & (9) 0.0126 & (9) 0.0064 & (9) 0.0171 & (9) 0.0087 & (9) 0.0187 & (9) 0.0095 & (6) 0.0004 & (5) 0.0004 & (5) 0.0004 & (3) 0.0004 & (7) 0.0197 & (7) 0.0116 & (11) 0.0240 & (11) 0.0136 & (4) 6.20 & (3) 5.80 & (4) 7.40 & (6) 6.80 \\
\midrule
\multirow{18}{*}{\rotatebox{90}{ArcFace~\cite{Deng2018ArcFaceAA}}} & FaceQnet~\cite{hernandez2019faceqnet} & (18) 0.0169 & (18) 0.0092 & (19) 0.0182 & (18) 0.0095 & (17) 0.0129 & (17) 0.0066 & (17) 0.0131 & (17) 0.0067 & (18) 0.0296 & (18) 0.0150 & (18) 0.0306 & (18) 0.0155 & (19) 0.0013 & (19) 0.0008 & (19) 0.0013 & (19) 0.0008 & (19) 0.0636 & (19) 0.0320 & (19) 0.0667 & (19) 0.0335 & (16) 18.20 & (15) 18.20 & (16) 18.40 & (13) 18.20 \\
 & SER-FIQA~\cite{Terhorst2020SERFIQUE} & (5) 0.0120 & (14) 0.0071 & (8) 0.0141 & (13) 0.0082 & (16) 0.0124 & (16) 0.0064 & (15) 0.0126 & (15) 0.0064 & (15) 0.0194 & (15) 0.0099 & (11) 0.0209 & (11) 0.0106 & (17) 0.0007 & (17) 0.0005 & (17) 0.0006 & (17) 0.0004 & (15) 0.0376 & (15) 0.0225 & (8) 0.0402 & (11) 0.0227 & (13) 13.60 & (13) 15.40 & (12) 11.80 & (11) 13.40 \\
 & PFE~\cite{shi2019probabilistic} & (10) 0.0123 & (3) 0.0064 & (5) 0.0139 & (3) 0.0071 & (12) 0.0117 & (12) 0.0060 & (7) 0.0122 & (7) 0.0063 & (16) 0.0195 & (16) 0.0099 & (17) 0.0228 & (17) 0.0116 & (7) 0.0004 & (8) 0.0004 & (8) 0.0004 & (6) 0.0004 & (3) 0.0299 & (2) \underline{0.0171} & (14) 0.0441 & (14) 0.0241 & (10) 9.60 & (7) 8.20 & (10) 10.20 & (6) 9.40 \\
 & MagFace~\cite{Meng2021MagFaceAU} & (6) 0.0120 & (5) 0.0065 & (6) 0.0140 & (5) 0.0074 & (5) 0.0114 & (6) 0.0059 & (5) 0.0120 & (5) 0.0062 & (14) 0.0194 & (14) 0.0098 & (14) 0.0224 & (14) 0.0114 & (5) 0.0004 & (5) 0.0004 & (7) 0.0004 & (7) 0.0004 & (18) 0.0579 & (18) 0.0301 & (18) 0.0601 & (18) 0.0308 & (10) 9.60 & (10) 9.60 & (9) 10.00 & (7) 9.80 \\
 & SDD-FIQA~\cite{Ou2021SDDFIQAUF} & (14) 0.0133 & (8) 0.0068 & (14) 0.0148 & (7) 0.0076 & (8) 0.0115 & (8) 0.0059 & (11) 0.0123 & (12) 0.0063 & (12) 0.0188 & (12) 0.0096 & (13) 0.0223 & (13) 0.0113 & (2) \underline{0.0004} & (2) \underline{0.0003} & (2) \underline{0.0004} & (2) \underline{0.0003} & (16) 0.0423 & (16) 0.0239 & (16) 0.0488 & (16) 0.0261 & (11) 10.40 & (9) 9.20 & (11) 11.20 & (8) 10.00 \\
 & LightQNet~\cite{9528058} & (8) 0.0122 & (4) 0.0064 & (7) 0.0140 & (4) 0.0073 & (11) 0.0116 & (11) 0.0060 & (12) 0.0123 & (11) 0.0063 & (13) 0.0194 & (13) 0.0098 & (16) 0.0228 & (16) 0.0116 & (16) 0.0006 & (16) 0.0005 & (6) 0.0004 & (8) 0.0004 & (10) 0.0358 & (10) 0.0211 & (10) 0.0405 & (7) 0.0223 & (12) 11.60 & (11) 10.80 & (10) 10.20 & (5) 9.20 \\
 & FaceQGen~\cite{HernandezOrtega2021FaceQgenSD} & (17) 0.0145 & (17) 0.0078 & (17) 0.0163 & (16) 0.0085 & (18) 0.0132 & (18) 0.0068 & (18) 0.0133 & (18) 0.0068 & (17) 0.0203 & (17) 0.0103 & (15) 0.0226 & (15) 0.0115 & (18) 0.0012 & (18) 0.0007 & (18) 0.0010 & (18) 0.0007 & (11) 0.0361 & (9) 0.0210 & (15) 0.0456 & (15) 0.0251 & (15) 16.20 & (14) 15.80 & (15) 16.60 & (12) 16.40 \\
 & FaceQAN~\cite{Babnik2022FaceQANFI} & (16) 0.0143 & (15) 0.0074 & (16) 0.0155 & (9) 0.0079 & (14) 0.0120 & (14) 0.0062 & (16) 0.0128 & (16) 0.0065 & (11) 0.0184 & (11) 0.0094 & (12) 0.0214 & (12) 0.0109 & (12) 0.0005 & (11) 0.0004 & (13) 0.0005 & (13) 0.0004 & (17) 0.0467 & (17) 0.0266 & (17) 0.0503 & (17) 0.0269 & (14) 14.00 & (12) 13.60 & (14) 14.80 & (11) 13.40 \\
 & CR-FIQA (L)~\cite{Boutros2021CRFIQAFI} & (2) \underline{0.0118} & (9) 0.0069 & (9) 0.0143 & (17) 0.0085 & (4) 0.0114 & (4) 0.0058 & (4) 0.0120 & (4) 0.0061 & (8) 0.0175 & (7) 0.0089 & (8) 0.0195 & (8) 0.0099 & (11) 0.0005 & (10) 0.0004 & (9) 0.0004 & (9) 0.0004 & (13) 0.0372 & (12) 0.0215 & (11) 0.0411 & (8) 0.0223 & (6) 7.60 & (8) 8.40 & (6) 8.20 & (5) 9.20 \\
 & DifFIQA~\cite{babnik2023diffiqa} & (15) 0.0135 & (16) 0.0075 & (15) 0.0153 & (15) 0.0084 & (9) 0.0115 & (9) 0.0059 & (8) 0.0122 & (8) 0.0063 & (7) 0.0175 & (8) 0.0089 & (7) 0.0192 & (7) 0.0098 & (8) 0.0005 & (7) 0.0004 & (12) 0.0005 & (12) 0.0004 & (8) 0.0340 & (8) 0.0202 & (6) 0.0392 & (5) 0.0217 & (9) 9.40 & (10) 9.60 & (7) 9.60 & (6) 9.40 \\
 & DifFIQA (R)~\cite{babnik2023diffiqa} & (13) 0.0127 & (7) 0.0068 & (12) 0.0146 & (8) 0.0076 & (10) 0.0115 & (10) 0.0059 & (6) 0.0121 & (6) 0.0062 & (3) 0.0165 & (3) 0.0084 & (4) 0.0186 & (4) 0.0095 & (15) 0.0006 & (14) 0.0004 & (15) 0.0005 & (15) 0.0004 & (5) 0.0322 & (6) 0.0196 & (2) \underline{0.0370} & (1) \textbf{0.0211} & (8) 9.20 & (6) 8.00 & (5) 7.80 & (4) 6.80 \\
 & CLIB-FIQA~\cite{Ou_2024_CVPR} & (9) 0.0122 & (10) 0.0070 & (10) 0.0143 & (10) 0.0080 & (3) 0.0110 & (3) 0.0057 & (3) 0.0119 & (3) 0.0061 & (4) 0.0165 & (4) 0.0084 & (3) 0.0185 & (3) 0.0094 & (4) 0.0004 & (3) 0.0004 & (3) 0.0004 & (3) 0.0004 & (4) 0.0304 & (4) 0.0184 & (1) \textbf{0.0370} & (2) \underline{0.0213} & (1) \textbf{4.80} & (2) \underline{4.80} & (2) \underline{4.00} & (1) \textbf{4.20} \\
 & GraFIQs~\cite{kolf2024grafiqs} & (12) 0.0125 & (13) 0.0071 & (11) 0.0145 & (12) 0.0082 & (15) 0.0121 & (15) 0.0062 & (14) 0.0125 & (14) 0.0064 & (10) 0.0181 & (10) 0.0093 & (10) 0.0208 & (10) 0.0106 & (14) 0.0006 & (15) 0.0004 & (16) 0.0005 & (16) 0.0004 & (1) \textbf{0.0274} & (1) \textbf{0.0164} & (12) 0.0413 & (12) 0.0232 & (11) 10.40 & (11) 10.80 & (13) 12.60 & (10) 12.80 \\
 & eDifFIQA (S)~\cite{10468647} & (11) 0.0124 & (11) 0.0071 & (13) 0.0147 & (14) 0.0083 & (6) 0.0114 & (5) 0.0058 & (9) 0.0122 & (9) 0.0063 & (6) 0.0170 & (6) 0.0086 & (6) 0.0191 & (6) 0.0097 & (9) 0.0005 & (9) 0.0004 & (14) 0.0005 & (14) 0.0004 & (9) 0.0356 & (11) 0.0212 & (7) 0.0402 & (10) 0.0225 & (7) 8.20 & (8) 8.40 & (8) 9.80 & (9) 10.60 \\
 & eDifFIQA (M)~\cite{10468647} & (4) 0.0120 & (12) 0.0071 & (4) 0.0138 & (11) 0.0082 & (1) \textbf{0.0103} & (1) \textbf{0.0053} & (1) \textbf{0.0111} & (1) \textbf{0.0057} & (1) \textbf{0.0163} & (1) \textbf{0.0083} & (1) \textbf{0.0184} & (1) \textbf{0.0094} & (10) 0.0005 & (12) 0.0004 & (10) 0.0004 & (10) 0.0004 & (12) 0.0364 & (13) 0.0216 & (4) 0.0380 & (4) 0.0214 & (4) 5.60 & (5) 7.80 & (2) \underline{4.00} & (3) 5.40 \\
 & eDifFIQA (L)~\cite{10468647} & (1) \textbf{0.0115} & (6) 0.0065 & (1) \textbf{0.0133} & (6) 0.0075 & (2) \underline{0.0109} & (2) \underline{0.0056} & (2) \underline{0.0118} & (2) \underline{0.0061} & (5) 0.0166 & (5) 0.0084 & (2) \underline{0.0185} & (2) \underline{0.0094} & (13) 0.0005 & (13) 0.0004 & (11) 0.0004 & (11) 0.0004 & (6) 0.0322 & (5) 0.0193 & (3) 0.0378 & (3) 0.0213 & (3) 5.40 & (4) 6.20 & (1) \textbf{3.80} & (2) \underline{4.80} \\
 & FROQ~\cite{Babnik2025FROQ1OF} & (7) 0.0121 & (2) \underline{0.0063} & (3) 0.0137 & (2) \underline{0.0070} & (13) 0.0119 & (13) 0.0061 & (13) 0.0123 & (13) 0.0063 & (2) \underline{0.0164} & (2) \underline{0.0084} & (5) 0.0188 & (5) 0.0096 & (1) \textbf{0.0003} & (1) \textbf{0.0003} & (1) \textbf{0.0003} & (1) \textbf{0.0003} & (2) \underline{0.0297} & (3) 0.0181 & (5) 0.0389 & (6) 0.0221 & (2) \underline{5.00} & (1) \textbf{4.20} & (3) 5.40 & (3) 5.40 \\
 & ViT-FIQA~\cite{atzori2025vit} & (3) 0.0119 & (1) \textbf{0.0062} & (2) \underline{0.0136} & (1) \textbf{0.0070} & (7) 0.0114 & (7) 0.0059 & (10) 0.0122 & (10) 0.0063 & (9) 0.0178 & (9) 0.0090 & (9) 0.0199 & (9) 0.0101 & (6) 0.0004 & (6) 0.0004 & (5) 0.0004 & (5) 0.0004 & (7) 0.0333 & (7) 0.0197 & (9) 0.0403 & (9) 0.0224 & (5) 6.40 & (3) 6.00 & (4) 7.00 & (4) 6.80 \\
\midrule
\multirow{18}{*}{\rotatebox{90}{MagFace~\cite{Meng2021MagFaceAU}}} & FaceQnet~\cite{hernandez2019faceqnet} & (18) 0.0197 & (18) 0.0108 & (19) 0.0214 & (18) 0.0112 & (17) 0.0130 & (17) 0.0067 & (17) 0.0134 & (17) 0.0068 & (18) 0.0356 & (18) 0.0181 & (18) 0.0377 & (18) 0.0191 & (19) 0.0013 & (19) 0.0008 & (19) 0.0013 & (19) 0.0008 & (19) 0.0814 & (19) 0.0404 & (19) 0.0862 & (19) 0.0428 & (17) 18.20 & (17) 18.20 & (16) 18.40 & (16) 18.20 \\
 & SER-FIQA~\cite{Terhorst2020SERFIQUE} & (3) 0.0131 & (15) 0.0087 & (6) 0.0161 & (16) 0.0100 & (16) 0.0127 & (16) 0.0065 & (15) 0.0129 & (15) 0.0066 & (13) 0.0222 & (13) 0.0115 & (10) 0.0251 & (10) 0.0130 & (17) 0.0007 & (17) 0.0005 & (17) 0.0006 & (17) 0.0004 & (15) 0.0483 & (15) 0.0327 & (7) 0.0537 & (11) 0.0330 & (14) 12.80 & (15) 15.20 & (11) 11.00 & (14) 13.80 \\
 & PFE~\cite{shi2019probabilistic} & (12) 0.0141 & (3) 0.0073 & (5) 0.0160 & (3) 0.0082 & (9) 0.0118 & (9) 0.0060 & (6) 0.0123 & (6) 0.0063 & (14) 0.0223 & (14) 0.0115 & (14) 0.0273 & (14) 0.0141 & (7) 0.0005 & (8) 0.0004 & (7) 0.0004 & (7) 0.0004 & (3) 0.0404 & (1) \textbf{0.0255} & (14) 0.0593 & (13) 0.0341 & (8) 9.00 & (5) 7.00 & (7) 9.20 & (6) 8.60 \\
 & MagFace~\cite{Meng2021MagFaceAU} & (7) 0.0137 & (5) 0.0075 & (8) 0.0162 & (6) 0.0087 & (4) 0.0114 & (4) 0.0059 & (4) 0.0121 & (4) 0.0062 & (15) 0.0225 & (15) 0.0117 & (15) 0.0274 & (15) 0.0142 & (5) 0.0004 & (5) 0.0004 & (8) 0.0004 & (6) 0.0004 & (18) 0.0755 & (18) 0.0393 & (18) 0.0792 & (18) 0.0401 & (11) 9.80 & (10) 9.40 & (10) 10.60 & (10) 9.80 \\
 & SDD-FIQA~\cite{Ou2021SDDFIQAUF} & (14) 0.0150 & (6) 0.0076 & (14) 0.0170 & (5) 0.0086 & (8) 0.0116 & (8) 0.0060 & (12) 0.0125 & (12) 0.0064 & (12) 0.0218 & (12) 0.0113 & (13) 0.0268 & (13) 0.0139 & (2) \underline{0.0004} & (2) \underline{0.0003} & (2) \underline{0.0004} & (2) \underline{0.0003} & (16) 0.0554 & (16) 0.0336 & (17) 0.0657 & (16) 0.0360 & (12) 10.40 & (8) 8.80 & (12) 11.60 & (9) 9.60 \\
 & LightQNet~\cite{9528058} & (10) 0.0139 & (4) 0.0074 & (7) 0.0162 & (4) 0.0084 & (10) 0.0118 & (10) 0.0060 & (10) 0.0125 & (11) 0.0064 & (16) 0.0229 & (16) 0.0118 & (16) 0.0285 & (16) 0.0147 & (16) 0.0006 & (16) 0.0005 & (6) 0.0004 & (8) 0.0004 & (11) 0.0465 & (11) 0.0313 & (10) 0.0547 & (7) 0.0327 & (13) 12.60 & (13) 11.40 & (8) 9.80 & (8) 9.20 \\
 & FaceQGen~\cite{HernandezOrtega2021FaceQgenSD} & (17) 0.0168 & (17) 0.0091 & (17) 0.0190 & (15) 0.0100 & (18) 0.0134 & (18) 0.0069 & (18) 0.0136 & (18) 0.0070 & (17) 0.0248 & (17) 0.0129 & (17) 0.0286 & (17) 0.0148 & (18) 0.0012 & (18) 0.0008 & (18) 0.0010 & (18) 0.0007 & (14) 0.0481 & (8) 0.0301 & (15) 0.0608 & (15) 0.0352 & (16) 16.80 & (16) 15.60 & (15) 17.00 & (15) 16.60 \\
 & FaceQAN~\cite{Babnik2022FaceQANFI} & (16) 0.0163 & (12) 0.0084 & (16) 0.0180 & (9) 0.0092 & (13) 0.0122 & (14) 0.0062 & (16) 0.0129 & (16) 0.0066 & (11) 0.0216 & (11) 0.0111 & (12) 0.0265 & (12) 0.0137 & (11) 0.0005 & (10) 0.0004 & (12) 0.0005 & (12) 0.0004 & (17) 0.0585 & (17) 0.0361 & (16) 0.0653 & (17) 0.0366 & (15) 13.60 & (14) 12.80 & (14) 14.40 & (13) 13.20 \\
 & CR-FIQA (L)~\cite{Boutros2021CRFIQAFI} & (5) 0.0133 & (11) 0.0084 & (10) 0.0165 & (17) 0.0105 & (5) 0.0115 & (5) 0.0059 & (5) 0.0121 & (5) 0.0062 & (7) 0.0200 & (7) 0.0103 & (7) 0.0236 & (7) 0.0123 & (10) 0.0005 & (11) 0.0004 & (9) 0.0004 & (9) 0.0004 & (12) 0.0475 & (12) 0.0315 & (11) 0.0547 & (6) 0.0326 & (7) 7.80 & (9) 9.20 & (6) 8.40 & (7) 8.80 \\
 & DifFIQA~\cite{babnik2023diffiqa} & (15) 0.0152 & (16) 0.0088 & (15) 0.0176 & (13) 0.0099 & (11) 0.0118 & (11) 0.0061 & (11) 0.0125 & (10) 0.0064 & (6) 0.0198 & (6) 0.0103 & (5) 0.0233 & (5) 0.0121 & (8) 0.0005 & (7) 0.0004 & (13) 0.0005 & (11) 0.0004 & (7) 0.0436 & (9) 0.0302 & (5) 0.0522 & (5) 0.0323 & (9) 9.40 & (11) 9.80 & (8) 9.80 & (7) 8.80 \\
 & DifFIQA (R)~\cite{babnik2023diffiqa} & (13) 0.0141 & (7) 0.0077 & (12) 0.0167 & (7) 0.0088 & (12) 0.0119 & (12) 0.0061 & (8) 0.0124 & (8) 0.0063 & (2) \underline{0.0191} & (2) \underline{0.0099} & (3) 0.0228 & (3) 0.0118 & (15) 0.0006 & (14) 0.0004 & (15) 0.0005 & (15) 0.0004 & (6) 0.0419 & (7) 0.0297 & (1) \textbf{0.0504} & (2) \underline{0.0318} & (10) 9.60 & (7) 8.40 & (5) 7.80 & (5) 7.00 \\
 & CLIB-FIQA~\cite{Ou_2024_CVPR} & (6) 0.0136 & (9) 0.0083 & (9) 0.0164 & (10) 0.0094 & (3) 0.0111 & (3) 0.0057 & (3) 0.0120 & (3) 0.0061 & (4) 0.0192 & (4) 0.0099 & (4) 0.0229 & (4) 0.0119 & (4) 0.0004 & (4) 0.0004 & (3) 0.0004 & (3) 0.0004 & (4) 0.0405 & (4) 0.0284 & (2) \underline{0.0510} & (4) 0.0321 & (1) \textbf{4.20} & (1) \textbf{4.80} & (2) \underline{4.20} & (1) \textbf{4.80} \\
 & GraFIQs~\cite{kolf2024grafiqs} & (9) 0.0138 & (10) 0.0084 & (11) 0.0166 & (11) 0.0096 & (15) 0.0124 & (15) 0.0063 & (14) 0.0128 & (14) 0.0065 & (10) 0.0211 & (10) 0.0110 & (11) 0.0251 & (11) 0.0131 & (14) 0.0006 & (15) 0.0004 & (16) 0.0006 & (16) 0.0004 & (1) \textbf{0.0368} & (2) \underline{0.0257} & (12) 0.0552 & (12) 0.0337 & (11) 9.80 & (12) 10.40 & (13) 12.80 & (12) 12.80 \\
 & eDifFIQA (S)~\cite{10468647} & (8) 0.0138 & (13) 0.0085 & (13) 0.0168 & (14) 0.0100 & (7) 0.0115 & (7) 0.0059 & (9) 0.0124 & (9) 0.0064 & (5) 0.0196 & (5) 0.0102 & (6) 0.0234 & (6) 0.0122 & (9) 0.0005 & (9) 0.0004 & (14) 0.0005 & (14) 0.0004 & (9) 0.0454 & (10) 0.0312 & (8) 0.0539 & (10) 0.0329 & (6) 7.60 & (8) 8.80 & (9) 10.00 & (11) 10.60 \\
 & eDifFIQA (M)~\cite{10468647} & (4) 0.0132 & (14) 0.0086 & (3) 0.0157 & (12) 0.0098 & (1) \textbf{0.0104} & (1) \textbf{0.0054} & (1) \textbf{0.0113} & (1) \textbf{0.0058} & (1) \textbf{0.0189} & (1) \textbf{0.0098} & (1) \textbf{0.0226} & (1) \textbf{0.0117} & (12) 0.0005 & (12) 0.0004 & (10) 0.0005 & (10) 0.0004 & (10) 0.0465 & (13) 0.0318 & (4) 0.0517 & (1) \textbf{0.0317} & (3) 5.60 & (6) 8.20 & (1) \textbf{3.80} & (2) \underline{5.00} \\
 & eDifFIQA (L)~\cite{10468647} & (1) \textbf{0.0127} & (8) 0.0078 & (1) \textbf{0.0152} & (8) 0.0088 & (2) \underline{0.0110} & (2) \underline{0.0057} & (2) \underline{0.0119} & (2) \underline{0.0061} & (3) 0.0192 & (3) 0.0099 & (2) \underline{0.0227} & (2) \underline{0.0118} & (13) 0.0005 & (13) 0.0004 & (11) 0.0005 & (13) 0.0004 & (5) 0.0419 & (5) 0.0291 & (3) 0.0514 & (3) 0.0319 & (2) \underline{4.80} & (4) 6.20 & (1) \textbf{3.80} & (3) 5.60 \\
 & FROQ~\cite{Babnik2025FROQ1OF} & (11) 0.0139 & (2) \underline{0.0072} & (4) 0.0158 & (2) \underline{0.0081} & (14) 0.0122 & (13) 0.0062 & (13) 0.0126 & (13) 0.0065 & (9) 0.0205 & (9) 0.0107 & (9) 0.0244 & (9) 0.0127 & (1) \textbf{0.0003} & (1) \textbf{0.0003} & (1) \textbf{0.0003} & (1) \textbf{0.0003} & (2) \underline{0.0399} & (3) 0.0278 & (6) 0.0529 & (9) 0.0328 & (5) 7.40 & (3) 5.60 & (4) 6.60 & (4) 6.80 \\
 & ViT-FIQA~\cite{atzori2025vit} & (2) \underline{0.0131} & (1) \textbf{0.0069} & (2) \underline{0.0154} & (1) \textbf{0.0079} & (6) 0.0115 & (6) 0.0059 & (7) 0.0124 & (7) 0.0063 & (8) 0.0201 & (8) 0.0104 & (8) 0.0238 & (8) 0.0124 & (6) 0.0004 & (6) 0.0004 & (5) 0.0004 & (4) 0.0004 & (8) 0.0441 & (6) 0.0297 & (9) 0.0547 & (8) 0.0327 & (4) 6.00 & (2) \underline{5.40} & (3) 6.20 & (3) 5.60 \\
\midrule
\multirow{18}{*}{\rotatebox{90}{SwinFace~\cite{qin2023swinface}}} & FaceQnet~\cite{hernandez2019faceqnet} & (19) 0.0211 & (18) 0.0119 & (19) 0.0230 & (18) 0.0123 & (17) 0.0131 & (17) 0.0067 & (17) 0.0135 & (17) 0.0069 & (18) 0.0320 & (18) 0.0161 & (18) 0.0333 & (18) 0.0168 & (19) 0.0013 & (19) 0.0008 & (19) 0.0013 & (19) 0.0008 & (19) 0.0612 & (19) 0.0307 & (19) 0.0636 & (19) 0.0319 & (18) 18.40 & (13) 18.20 & (17) 18.40 & (15) 18.20 \\
 & SER-FIQA~\cite{Terhorst2020SERFIQUE} & (1) \textbf{0.0132} & (14) 0.0098 & (2) \underline{0.0164} & (11) 0.0110 & (16) 0.0128 & (16) 0.0065 & (15) 0.0129 & (15) 0.0066 & (15) 0.0210 & (15) 0.0107 & (11) 0.0230 & (10) 0.0117 & (17) 0.0007 & (17) 0.0005 & (17) 0.0006 & (17) 0.0004 & (14) 0.0318 & (15) 0.0199 & (7) 0.0349 & (8) 0.0201 & (14) 12.60 & (11) 15.40 & (11) 10.40 & (11) 12.20 \\
 & PFE~\cite{shi2019probabilistic} & (11) 0.0156 & (3) 0.0083 & (10) 0.0181 & (3) 0.0093 & (12) 0.0120 & (12) 0.0061 & (8) 0.0125 & (8) 0.0064 & (16) 0.0210 & (16) 0.0107 & (16) 0.0250 & (16) 0.0127 & (7) 0.0004 & (8) 0.0004 & (8) 0.0004 & (7) 0.0004 & (4) 0.0272 & (2) \underline{0.0159} & (15) 0.0404 & (14) 0.0224 & (11) 10.00 & (5) 8.20 & (13) 11.40 & (9) 9.60 \\
 & MagFace~\cite{Meng2021MagFaceAU} & (9) 0.0147 & (6) 0.0086 & (7) 0.0179 & (6) 0.0100 & (6) 0.0117 & (6) 0.0060 & (5) 0.0123 & (5) 0.0063 & (13) 0.0208 & (13) 0.0106 & (13) 0.0246 & (14) 0.0126 & (5) 0.0004 & (4) 0.0004 & (7) 0.0004 & (5) 0.0004 & (18) 0.0545 & (18) 0.0284 & (18) 0.0565 & (18) 0.0289 & (12) 10.20 & (7) 9.40 & (10) 10.00 & (9) 9.60 \\
 & SDD-FIQA~\cite{Ou2021SDDFIQAUF} & (15) 0.0168 & (5) 0.0086 & (15) 0.0191 & (5) 0.0096 & (8) 0.0118 & (8) 0.0060 & (12) 0.0127 & (12) 0.0065 & (12) 0.0206 & (12) 0.0105 & (14) 0.0246 & (13) 0.0125 & (2) \underline{0.0004} & (2) \underline{0.0003} & (2) \underline{0.0004} & (2) \underline{0.0003} & (16) 0.0388 & (16) 0.0224 & (17) 0.0450 & (17) 0.0243 & (13) 10.60 & (6) 8.60 & (14) 12.00 & (10) 9.80 \\
 & LightQNet~\cite{9528058} & (10) 0.0153 & (4) 0.0084 & (12) 0.0185 & (4) 0.0096 & (11) 0.0119 & (11) 0.0061 & (11) 0.0127 & (11) 0.0065 & (14) 0.0209 & (14) 0.0107 & (17) 0.0251 & (17) 0.0128 & (16) 0.0006 & (16) 0.0005 & (6) 0.0004 & (8) 0.0004 & (13) 0.0314 & (11) 0.0191 & (10) 0.0360 & (9) 0.0201 & (15) 12.80 & (9) 11.20 & (12) 11.20 & (10) 9.80 \\
 & FaceQGen~\cite{HernandezOrtega2021FaceQgenSD} & (16) 0.0183 & (17) 0.0103 & (17) 0.0210 & (13) 0.0111 & (18) 0.0134 & (18) 0.0069 & (18) 0.0136 & (18) 0.0070 & (17) 0.0221 & (17) 0.0113 & (15) 0.0250 & (15) 0.0127 & (18) 0.0012 & (18) 0.0007 & (18) 0.0010 & (18) 0.0007 & (11) 0.0312 & (9) 0.0189 & (14) 0.0399 & (15) 0.0226 & (17) 16.00 & (12) 15.80 & (16) 16.40 & (14) 15.80 \\
 & FaceQAN~\cite{Babnik2022FaceQANFI} & (17) 0.0184 & (9) 0.0096 & (16) 0.0203 & (9) 0.0104 & (14) 0.0124 & (14) 0.0063 & (16) 0.0131 & (16) 0.0067 & (11) 0.0201 & (11) 0.0102 & (12) 0.0239 & (12) 0.0122 & (11) 0.0005 & (11) 0.0004 & (13) 0.0005 & (13) 0.0004 & (17) 0.0410 & (17) 0.0239 & (16) 0.0447 & (16) 0.0243 & (16) 14.00 & (10) 12.40 & (15) 14.60 & (12) 13.20 \\
 & CR-FIQA (L)~\cite{Boutros2021CRFIQAFI} & (3) 0.0134 & (10) 0.0096 & (1) \textbf{0.0163} & (17) 0.0116 & (5) 0.0117 & (5) 0.0060 & (4) 0.0123 & (4) 0.0063 & (8) 0.0188 & (8) 0.0096 & (8) 0.0216 & (8) 0.0110 & (12) 0.0005 & (12) 0.0004 & (9) 0.0004 & (9) 0.0004 & (12) 0.0314 & (12) 0.0191 & (9) 0.0354 & (7) 0.0199 & (7) 8.00 & (7) 9.40 & (4) 6.20 & (7) 9.00 \\
 & DifFIQA~\cite{babnik2023diffiqa} & (14) 0.0162 & (15) 0.0099 & (14) 0.0189 & (15) 0.0111 & (10) 0.0118 & (10) 0.0060 & (10) 0.0125 & (10) 0.0064 & (7) 0.0187 & (7) 0.0095 & (5) 0.0211 & (5) 0.0108 & (8) 0.0005 & (7) 0.0004 & (12) 0.0005 & (11) 0.0004 & (7) 0.0292 & (8) 0.0181 & (5) 0.0342 & (5) 0.0195 & (9) 9.20 & (7) 9.40 & (9) 9.20 & (8) 9.20 \\
 & DifFIQA (R)~\cite{babnik2023diffiqa} & (12) 0.0158 & (7) 0.0089 & (13) 0.0189 & (7) 0.0100 & (9) 0.0118 & (9) 0.0060 & (6) 0.0124 & (6) 0.0063 & (3) 0.0179 & (3) 0.0091 & (4) 0.0208 & (3) 0.0106 & (15) 0.0006 & (14) 0.0004 & (15) 0.0005 & (15) 0.0004 & (5) 0.0274 & (6) 0.0173 & (1) \textbf{0.0324} & (1) \textbf{0.0189} & (8) 8.80 & (3) 7.80 & (6) 7.80 & (5) 6.40 \\
 & CLIB-FIQA~\cite{Ou_2024_CVPR} & (5) 0.0143 & (11) 0.0096 & (6) 0.0177 & (10) 0.0108 & (3) 0.0112 & (3) 0.0058 & (3) 0.0121 & (3) 0.0062 & (4) 0.0179 & (4) 0.0091 & (3) 0.0208 & (4) 0.0106 & (4) 0.0004 & (3) 0.0004 & (3) 0.0004 & (3) 0.0004 & (3) 0.0265 & (4) 0.0164 & (2) \underline{0.0330} & (4) 0.0191 & (1) \textbf{3.80} & (1) \textbf{5.00} & (1) \textbf{3.40} & (1) \textbf{4.80} \\
 & GraFIQs~\cite{kolf2024grafiqs} & (8) 0.0146 & (13) 0.0097 & (9) 0.0181 & (14) 0.0111 & (15) 0.0124 & (15) 0.0064 & (14) 0.0128 & (14) 0.0066 & (10) 0.0197 & (10) 0.0101 & (10) 0.0230 & (11) 0.0117 & (14) 0.0006 & (15) 0.0004 & (16) 0.0005 & (16) 0.0004 & (1) \textbf{0.0235} & (1) \textbf{0.0146} & (11) 0.0363 & (12) 0.0210 & (10) 9.60 & (8) 10.80 & (14) 12.00 & (13) 13.40 \\
 & eDifFIQA (S)~\cite{10468647} & (6) 0.0143 & (12) 0.0097 & (8) 0.0179 & (12) 0.0111 & (4) 0.0115 & (4) 0.0059 & (7) 0.0125 & (7) 0.0064 & (5) 0.0183 & (5) 0.0094 & (6) 0.0212 & (6) 0.0108 & (9) 0.0005 & (9) 0.0004 & (14) 0.0005 & (14) 0.0004 & (9) 0.0304 & (10) 0.0190 & (8) 0.0354 & (10) 0.0203 & (4) 6.60 & (4) 8.00 & (8) 8.60 & (10) 9.80 \\
 & eDifFIQA (M)~\cite{10468647} & (4) 0.0134 & (16) 0.0100 & (3) 0.0164 & (16) 0.0111 & (1) \textbf{0.0105} & (1) \textbf{0.0054} & (1) \textbf{0.0114} & (1) \textbf{0.0058} & (1) \textbf{0.0175} & (1) \textbf{0.0089} & (1) \textbf{0.0203} & (1) \textbf{0.0103} & (10) 0.0005 & (10) 0.0004 & (10) 0.0004 & (10) 0.0004 & (10) 0.0309 & (13) 0.0192 & (3) 0.0330 & (2) \underline{0.0189} & (3) 5.20 & (5) 8.20 & (2) \underline{3.60} & (4) 6.00 \\
 & eDifFIQA (L)~\cite{10468647} & (2) \underline{0.0133} & (8) 0.0090 & (4) 0.0164 & (8) 0.0102 & (2) \underline{0.0112} & (2) \underline{0.0057} & (2) \underline{0.0121} & (2) \underline{0.0062} & (2) \underline{0.0178} & (2) \underline{0.0091} & (2) \underline{0.0205} & (2) \underline{0.0104} & (13) 0.0005 & (13) 0.0004 & (11) 0.0004 & (12) 0.0004 & (6) 0.0277 & (5) 0.0173 & (4) 0.0333 & (3) 0.0191 & (2) \underline{5.00} & (2) \underline{6.00} & (3) 4.60 & (2) \underline{5.40} \\
 & FROQ~\cite{Babnik2025FROQ1OF} & (13) 0.0159 & (2) \underline{0.0082} & (11) 0.0182 & (2) \underline{0.0093} & (13) 0.0122 & (13) 0.0063 & (13) 0.0127 & (13) 0.0065 & (6) 0.0186 & (6) 0.0095 & (7) 0.0216 & (7) 0.0110 & (1) \textbf{0.0003} & (1) \textbf{0.0003} & (1) \textbf{0.0003} & (1) \textbf{0.0003} & (2) \underline{0.0255} & (3) 0.0160 & (6) 0.0343 & (6) 0.0198 & (5) 7.00 & (1) \textbf{5.00} & (5) 7.60 & (3) 5.80 \\
 & ViT-FIQA~\cite{atzori2025vit} & (7) 0.0145 & (1) \textbf{0.0079} & (5) 0.0174 & (1) \textbf{0.0091} & (7) 0.0117 & (7) 0.0060 & (9) 0.0125 & (9) 0.0064 & (9) 0.0190 & (9) 0.0097 & (9) 0.0218 & (9) 0.0111 & (6) 0.0004 & (6) 0.0004 & (5) 0.0004 & (4) 0.0004 & (8) 0.0296 & (7) 0.0180 & (12) 0.0367 & (11) 0.0205 & (6) 7.40 & (2) \underline{6.00} & (7) 8.00 & (6) 6.80 \\
\bottomrule
\end{tabular}%
}

\caption{\textbf{Comparison of FIQA Methods Across Different EDC-Based Variants.} 18 state-of-the-art FIQA solutions are compared on 4 FR models. The performance is measured using pAUC@0.3, computed on FNMR for EDC and PPD-EDC, and on CER for FT-EDC and FT-PPD-EDC protocols. Values in parentheses denote ranks (lower is better). The Summary block reports the average rank across datasets (lower is better). Best and second-best results are highlighted. FIQA methods across the EDC protocols remain stable.}
\label{tab:fiqa_pauc}
\end{sidewaystable*}



\begin{table*}[t]
\vspace{-1.5mm}
\centering
\scriptsize
\setlength{\tabcolsep}{3pt}
\renewcommand{\arraystretch}{0.95}
\begin{adjustbox}{max totalsize={\textwidth}{0.82\textheight},center}
\begin{tabular}{ll|r|r|r|r|r|r|r|r|r|r|r|r}
\toprule
\multirow{3}{*}{FR} & \multirow{3}{*}{FIQA} & \multicolumn{2}{c|}{Adience~\cite{6906255}} & \multicolumn{2}{c|}{CALFW~\cite{calfw}} & \multicolumn{2}{c|}{CPLFW~\cite{cplfw}} & \multicolumn{2}{c|}{LFW~\cite{huang:inria-00321923}} & \multicolumn{2}{c|}{XQLFW~\cite{knoche2021cross}} & \multicolumn{2}{c}{Summary} \\
 &  & \multicolumn{2}{c|}{RCE} & \multicolumn{2}{c|}{RCE} & \multicolumn{2}{c|}{RCE} & \multicolumn{2}{c|}{RCE} & \multicolumn{2}{c|}{RCE} & \multicolumn{2}{c|}{RCE Avg. Rank} \\
 &  & $\alpha=0$ & $\alpha=3$ & $\alpha=0$ & $\alpha=3$ & $\alpha=0$ & $\alpha=3$ & $\alpha=0$ & $\alpha=3$ & $\alpha=0$ & $\alpha=3$ & $\alpha=0$ & $\alpha=3$ \\
\midrule
\multirow{18}{*}{\rotatebox{90}{AdaFace~\cite{Kim2022AdaFaceQA}}} & FaceQnet~\cite{hernandez2019faceqnet} & (19) 0.1567 & (19) -0.2040 & (18) 0.3071 & (18) -0.2243 & (18) 0.3074 & (18) -0.1837 & (13) 0.2966 & (15) -0.3566 & (18) -0.0369 & (19) -0.4971 & (14) 17.20 & (16) 17.80 \\
 & SER-FIQA~\cite{Terhorst2020SERFIQUE} & (1) \textbf{0.3534} & (1) \textbf{0.1773} & (17) 0.3102 & (17) -0.2149 & (17) 0.3497 & (17) -0.0747 & (9) 0.3049 & (6) -0.2590 & (3) 0.0178 & (3) -0.3919 & (6) 9.40 & (6) 8.80 \\
 & PFE~\cite{shi2019probabilistic} & (12) 0.2251 & (12) 0.0030 & (8) 0.3350 & (9) -0.1182 & (14) 0.3612 & (12) -0.0196 & (17) 0.2858 & (16) -0.3818 & (10) -0.0013 & (8) -0.4276 & (12) 12.20 & (11) 11.40 \\
 & MagFace~\cite{Meng2021MagFaceAU} & (10) 0.2519 & (6) 0.0906 & (4) 0.3392 & (4) -0.1100 & (13) 0.3632 & (11) -0.0127 & (11) 0.2996 & (13) -0.3211 & (19) -0.0594 & (18) -0.4879 & (11) 11.40 & (10) 10.40 \\
 & SDD-FIQA~\cite{Ou2021SDDFIQAUF} & (16) 0.1816 & (13) -0.0472 & (7) 0.3361 & (7) -0.1159 & (10) 0.3644 & (9) -0.0057 & (7) 0.3071 & (9) -0.2871 & (16) -0.0194 & (12) -0.4530 & (10) 11.20 & (9) 10.00 \\
 & LightQNet~\cite{9528058} & (7) 0.2725 & (5) 0.1122 & (5) 0.3368 & (6) -0.1134 & (15) 0.3581 & (15) -0.0323 & (15) 0.2947 & (10) -0.3114 & (12) -0.0068 & (13) -0.4596 & (9) 10.80 & (8) 9.80 \\
 & FaceQGen~\cite{HernandezOrtega2021FaceQgenSD} & (11) 0.2451 & (10) 0.0238 & (16) 0.3137 & (16) -0.1992 & (16) 0.3554 & (16) -0.0337 & (3) 0.3148 & (3) -0.2270 & (1) \textbf{0.0237} & (2) \underline{-0.3769} & (6) 9.40 & (7) 9.40 \\
 & FaceQAN~\cite{Babnik2022FaceQANFI} & (15) 0.1968 & (14) -0.0514 & (14) 0.3265 & (14) -0.1502 & (9) 0.3669 & (8) -0.0050 & (10) 0.3012 & (11) -0.3131 & (15) -0.0157 & (14) -0.4596 & (13) 12.60 & (15) 12.20 \\
 & CR-FIQA (L)~\cite{Boutros2021CRFIQAFI} & (6) 0.2757 & (11) 0.0093 & (11) 0.3330 & (11) -0.1350 & (8) 0.3690 & (10) -0.0107 & (12) 0.2986 & (14) -0.3238 & (13) -0.0128 & (15) -0.4616 & (7) 10.00 & (15) 12.20 \\
 & DifFIQA~\cite{babnik2023diffiqa} & (13) 0.2236 & (15) -0.0518 & (9) 0.3343 & (8) -0.1169 & (7) 0.3714 & (6) 0.0061 & (5) 0.3122 & (4) -0.2526 & (4) 0.0124 & (4) -0.4044 & (4) 7.60 & (5) 7.40 \\
 & DifFIQA (R)~\cite{babnik2023diffiqa} & (14) 0.2145 & (17) -0.0632 & (13) 0.3288 & (12) -0.1378 & (4) 0.3767 & (4) 0.0210 & (18) 0.2814 & (18) -0.3973 & (7) 0.0062 & (9) -0.4338 & (10) 11.20 & (14) 12.00 \\
 & CLIB-FIQA~\cite{Ou_2024_CVPR} & (4) 0.3295 & (4) 0.1214 & (2) \underline{0.3396} & (2) \underline{-0.1011} & (3) 0.3791 & (3) 0.0304 & (14) 0.2950 & (12) -0.3204 & (5) 0.0098 & (6) -0.4174 & (3) 5.60 & (4) 5.40 \\
 & GraFIQs~\cite{kolf2024grafiqs} & (8) 0.2724 & (9) 0.0730 & (15) 0.3210 & (15) -0.1718 & (12) 0.3637 & (13) -0.0209 & (4) 0.3124 & (8) -0.2752 & (6) 0.0079 & (7) -0.4202 & (5) 9.00 & (10) 10.40 \\
 & eDifFIQA (S)~\cite{10468647} & (5) 0.3102 & (8) 0.0786 & (6) 0.3364 & (5) -0.1101 & (5) 0.3751 & (5) 0.0178 & (2) \underline{0.3178} & (2) \underline{-0.2184} & (9) -0.0003 & (5) -0.4123 & (2) \underline{5.40} & (3) 5.00 \\
 & eDifFIQA (M)~\cite{10468647} & (2) \underline{0.3461} & (2) \underline{0.1530} & (1) \textbf{0.3466} & (1) \textbf{-0.0736} & (2) \underline{0.3818} & (2) \underline{0.0417} & (8) 0.3053 & (5) -0.2551 & (14) -0.0130 & (11) -0.4496 & (2) \underline{5.40} & (2) \underline{4.20} \\
 & eDifFIQA (L)~\cite{10468647} & (3) 0.3406 & (3) 0.1524 & (3) 0.3393 & (3) -0.1059 & (1) \textbf{0.3841} & (1) \textbf{0.0519} & (1) \textbf{0.3254} & (1) \textbf{-0.1688} & (2) \underline{0.0226} & (1) \textbf{-0.3663} & (1) \textbf{2.00} & (1) \textbf{1.80} \\
 & FROQ~\cite{Babnik2025FROQ1OF} & (17) 0.1757 & (16) -0.0627 & (12) 0.3299 & (13) -0.1427 & (6) 0.3724 & (7) 0.0009 & (6) 0.3090 & (7) -0.2750 & (11) -0.0052 & (16) -0.4709 & (8) 10.40 & (13) 11.80 \\
 & ViT-FIQA~\cite{atzori2025vit} & (9) 0.2552 & (7) 0.0816 & (10) 0.3341 & (10) -0.1274 & (11) 0.3639 & (14) -0.0243 & (16) 0.2879 & (17) -0.3844 & (8) 0.0013 & (10) -0.4451 & (9) 10.80 & (12) 11.60 \\
\midrule
\multirow{18}{*}{\rotatebox{90}{ArcFace~\cite{Deng2018ArcFaceAA}}} & FaceQnet~\cite{hernandez2019faceqnet} & (18) 0.1914 & (18) -0.0582 & (18) 0.3075 & (18) -0.2236 & (18) 0.2969 & (18) -0.1708 & (15) 0.2959 & (16) -0.3626 & (15) -0.0481 & (16) -0.4355 & (17) 16.80 & (17) 17.20 \\
 & SER-FIQA~\cite{Terhorst2020SERFIQUE} & (1) \textbf{0.4056} & (1) \textbf{0.3275} & (17) 0.3138 & (17) -0.2042 & (17) 0.3463 & (17) -0.0587 & (6) 0.3123 & (4) -0.2296 & (3) -0.0126 & (3) -0.3565 & (7) 8.80 & (6) 8.40 \\
 & PFE~\cite{shi2019probabilistic} & (12) 0.2629 & (11) 0.1639 & (10) 0.3379 & (8) -0.1051 & (15) 0.3655 & (14) 0.0149 & (18) 0.2884 & (17) -0.3725 & (9) -0.0310 & (8) -0.3902 & (14) 12.80 & (13) 11.60 \\
 & MagFace~\cite{Meng2021MagFaceAU} & (9) 0.3016 & (5) 0.2635 & (2) \underline{0.3499} & (4) -0.0690 & (11) 0.3703 & (12) 0.0291 & (11) 0.3027 & (11) -0.3096 & (19) -0.0949 & (19) -0.4528 & (12) 10.40 & (9) 10.20 \\
 & SDD-FIQA~\cite{Ou2021SDDFIQAUF} & (16) 0.2128 & (14) 0.0851 & (12) 0.3358 & (11) -0.1153 & (13) 0.3693 & (8) 0.0320 & (9) 0.3099 & (9) -0.2817 & (17) -0.0551 & (11) -0.4173 & (15) 13.40 & (10) 10.60 \\
 & LightQNet~\cite{9528058} & (8) 0.3201 & (6) 0.2601 & (13) 0.3344 & (13) -0.1216 & (16) 0.3603 & (16) 0.0023 & (16) 0.2943 & (13) -0.3177 & (11) -0.0399 & (13) -0.4229 & (14) 12.80 & (14) 12.20 \\
 & FaceQGen~\cite{HernandezOrtega2021FaceQgenSD} & (11) 0.2680 & (12) 0.1393 & (16) 0.3204 & (16) -0.1738 & (14) 0.3658 & (9) 0.0320 & (7) 0.3121 & (6) -0.2419 & (1) \textbf{-0.0077} & (2) \underline{-0.3468} & (11) 9.80 & (7) 9.00 \\
 & FaceQAN~\cite{Babnik2022FaceQANFI} & (15) 0.2239 & (13) 0.0933 & (15) 0.3302 & (14) -0.1374 & (9) 0.3737 & (7) 0.0327 & (13) 0.2997 & (14) -0.3207 & (16) -0.0510 & (15) -0.4300 & (16) 13.60 & (15) 12.60 \\
 & CR-FIQA (L)~\cite{Boutros2021CRFIQAFI} & (6) 0.3307 & (10) 0.1934 & (6) 0.3412 & (7) -0.0926 & (8) 0.3754 & (11) 0.0297 & (12) 0.3017 & (12) -0.3138 & (13) -0.0461 & (14) -0.4275 & (8) 9.00 & (11) 10.80 \\
 & DifFIQA~\cite{babnik2023diffiqa} & (13) 0.2412 & (15) 0.0712 & (8) 0.3389 & (10) -0.1076 & (6) 0.3789 & (6) 0.0461 & (3) 0.3240 & (3) -0.2038 & (4) -0.0234 & (4) -0.3751 & (5) 6.80 & (5) 7.60 \\
 & DifFIQA (R)~\cite{babnik2023diffiqa} & (14) 0.2333 & (17) 0.0502 & (9) 0.3379 & (9) -0.1072 & (4) 0.3870 & (4) 0.0679 & (14) 0.2971 & (15) -0.3321 & (6) -0.0287 & (9) -0.4010 & (10) 9.40 & (11) 10.80 \\
 & CLIB-FIQA~\cite{Ou_2024_CVPR} & (4) 0.3811 & (4) 0.2727 & (4) 0.3487 & (2) \underline{-0.0597} & (3) 0.3872 & (3) 0.0697 & (10) 0.3037 & (10) -0.2872 & (5) -0.0239 & (5) -0.3815 & (2) \underline{5.20} & (3) 4.80 \\
 & GraFIQs~\cite{kolf2024grafiqs} & (7) 0.3223 & (7) 0.2377 & (14) 0.3314 & (15) -0.1425 & (10) 0.3710 & (13) 0.0286 & (4) 0.3162 & (8) -0.2608 & (7) -0.0292 & (7) -0.3893 & (6) 8.40 & (8) 10.00 \\
 & eDifFIQA (S)~\cite{10468647} & (5) 0.3598 & (8) 0.2375 & (7) 0.3408 & (5) -0.0904 & (5) 0.3826 & (5) 0.0524 & (2) \underline{0.3246} & (2) \underline{-0.1933} & (10) -0.0362 & (6) -0.3831 & (4) 5.80 & (4) 5.20 \\
 & eDifFIQA (M)~\cite{10468647} & (2) \underline{0.3984} & (3) 0.3067 & (1) \textbf{0.3517} & (1) \textbf{-0.0479} & (2) \underline{0.3910} & (2) \underline{0.0846} & (8) 0.3120 & (5) -0.2319 & (14) -0.0468 & (12) -0.4176 & (3) 5.40 & (2) \underline{4.60} \\
 & eDifFIQA (L)~\cite{10468647} & (3) 0.3964 & (2) \underline{0.3102} & (3) 0.3489 & (3) -0.0648 & (1) \textbf{0.3930} & (1) \textbf{0.0928} & (1) \textbf{0.3346} & (1) \textbf{-0.1323} & (2) \underline{-0.0094} & (1) \textbf{-0.3331} & (1) \textbf{2.00} & (1) \textbf{1.60} \\
 & FROQ~\cite{Babnik2025FROQ1OF} & (17) 0.2024 & (16) 0.0628 & (5) 0.3423 & (6) -0.0925 & (7) 0.3762 & (10) 0.0313 & (5) 0.3135 & (7) -0.2570 & (12) -0.0422 & (17) -0.4374 & (9) 9.20 & (12) 11.20 \\
 & ViT-FIQA~\cite{atzori2025vit} & (10) 0.2970 & (9) 0.2324 & (11) 0.3374 & (12) -0.1156 & (12) 0.3699 & (15) 0.0144 & (17) 0.2892 & (18) -0.3851 & (8) -0.0295 & (10) -0.4087 & (13) 11.60 & (16) 12.80 \\
\midrule
\multirow{18}{*}{\rotatebox{90}{MagFace~\cite{Meng2021MagFaceAU}}} & FaceQnet~\cite{hernandez2019faceqnet} & (18) 0.2057 & (18) -0.0137 & (18) 0.3037 & (18) -0.2352 & (18) 0.2924 & (18) -0.1051 & (14) 0.2917 & (16) -0.3689 & (13) -0.0766 & (14) -0.4089 & (15) 16.20 & (15) 16.80 \\
 & SER-FIQA~\cite{Terhorst2020SERFIQUE} & (2) \underline{0.4087} & (1) \textbf{0.3459} & (16) 0.3231 & (16) -0.1679 & (17) 0.3523 & (17) 0.0248 & (6) 0.3076 & (3) -0.2402 & (2) \underline{-0.0391} & (3) -0.3357 & (5) 8.60 & (6) 8.00 \\
 & PFE~\cite{shi2019probabilistic} & (11) 0.2825 & (11) 0.2125 & (10) 0.3333 & (9) -0.1215 & (14) 0.3728 & (13) 0.0933 & (18) 0.2820 & (17) -0.3901 & (8) -0.0602 & (8) -0.3699 & (13) 12.20 & (12) 11.60 \\
 & MagFace~\cite{Meng2021MagFaceAU} & (9) 0.3229 & (4) 0.3111 & (4) 0.3420 & (5) -0.0971 & (11) 0.3763 & (10) 0.1066 & (12) 0.2959 & (14) -0.3309 & (19) -0.1270 & (19) -0.4302 & (10) 11.00 & (8) 10.40 \\
 & SDD-FIQA~\cite{Ou2021SDDFIQAUF} & (16) 0.2319 & (13) 0.1457 & (14) 0.3303 & (14) -0.1371 & (12) 0.3763 & (9) 0.1072 & (9) 0.3045 & (9) -0.2947 & (17) -0.0824 & (11) -0.3961 & (14) 13.60 & (11) 11.20 \\
 & LightQNet~\cite{9528058} & (7) 0.3364 & (5) 0.3055 & (13) 0.3309 & (13) -0.1322 & (15) 0.3660 & (15) 0.0836 & (15) 0.2888 & (13) -0.3303 & (11) -0.0697 & (13) -0.4052 & (13) 12.20 & (13) 11.80 \\
 & FaceQGen~\cite{HernandezOrtega2021FaceQgenSD} & (12) 0.2792 & (12) 0.1758 & (17) 0.3138 & (17) -0.1909 & (16) 0.3616 & (16) 0.0811 & (7) 0.3066 & (6) -0.2591 & (1) \textbf{-0.0332} & (2) \underline{-0.3322} & (9) 10.60 & (9) 10.60 \\
 & FaceQAN~\cite{Babnik2022FaceQANFI} & (14) 0.2387 & (14) 0.1402 & (8) 0.3344 & (8) -0.1139 & (8) 0.3849 & (7) 0.1216 & (10) 0.2982 & (11) -0.3213 & (16) -0.0822 & (16) -0.4122 & (11) 11.20 & (11) 11.20 \\
 & CR-FIQA (L)~\cite{Boutros2021CRFIQAFI} & (6) 0.3475 & (10) 0.2329 & (11) 0.3331 & (11) -0.1223 & (7) 0.3876 & (8) 0.1179 & (11) 0.2972 & (12) -0.3256 & (14) -0.0799 & (17) -0.4141 & (7) 9.80 & (12) 11.60 \\
 & DifFIQA~\cite{babnik2023diffiqa} & (13) 0.2411 & (16) 0.0947 & (3) 0.3428 & (3) -0.0889 & (6) 0.3883 & (6) 0.1221 & (3) 0.3143 & (4) -0.2405 & (6) -0.0557 & (4) -0.3546 & (4) 6.20 & (5) 6.60 \\
 & DifFIQA (R)~\cite{babnik2023diffiqa} & (15) 0.2359 & (17) 0.0832 & (9) 0.3341 & (10) -0.1216 & (3) 0.3999 & (3) 0.1531 & (16) 0.2872 & (15) -0.3645 & (7) -0.0590 & (9) -0.3787 & (8) 10.00 & (10) 10.80 \\
 & CLIB-FIQA~\cite{Ou_2024_CVPR} & (4) 0.3924 & (6) 0.3044 & (5) 0.3411 & (4) -0.0905 & (4) 0.3952 & (4) 0.1435 & (13) 0.2954 & (10) -0.3133 & (5) -0.0550 & (6) -0.3596 & (4) 6.20 & (4) 6.00 \\
 & GraFIQs~\cite{kolf2024grafiqs} & (8) 0.3359 & (9) 0.2735 & (15) 0.3298 & (15) -0.1432 & (13) 0.3733 & (12) 0.0972 & (4) 0.3107 & (8) -0.2768 & (4) -0.0540 & (5) -0.3577 & (6) 8.80 & (7) 9.80 \\
 & eDifFIQA (S)~\cite{10468647} & (5) 0.3734 & (8) 0.2776 & (7) 0.3382 & (7) -0.1020 & (5) 0.3923 & (5) 0.1313 & (2) \underline{0.3163} & (2) \underline{-0.2220} & (10) -0.0688 & (7) -0.3640 & (3) 5.80 & (3) 5.80 \\
 & eDifFIQA (M)~\cite{10468647} & (1) \textbf{0.4098} & (3) 0.3393 & (1) \textbf{0.3515} & (1) \textbf{-0.0502} & (2) \underline{0.4025} & (2) \underline{0.1654} & (8) 0.3046 & (5) -0.2574 & (15) -0.0812 & (12) -0.3996 & (2) \underline{5.40} & (2) \underline{4.60} \\
 & eDifFIQA (L)~\cite{10468647} & (3) 0.4082 & (2) \underline{0.3420} & (2) \underline{0.3432} & (2) \underline{-0.0856} & (1) \textbf{0.4042} & (1) \textbf{0.1706} & (1) \textbf{0.3250} & (1) \textbf{-0.1680} & (3) -0.0418 & (1) \textbf{-0.3104} & (1) \textbf{2.00} & (1) \textbf{1.40} \\
 & FROQ~\cite{Babnik2025FROQ1OF} & (17) 0.2177 & (15) 0.1232 & (6) 0.3390 & (6) -0.0995 & (9) 0.3767 & (11) 0.0993 & (5) 0.3076 & (7) -0.2759 & (12) -0.0703 & (15) -0.4115 & (7) 9.80 & (10) 10.80 \\
 & ViT-FIQA~\cite{atzori2025vit} & (10) 0.3199 & (7) 0.2832 & (12) 0.3331 & (12) -0.1313 & (10) 0.3765 & (14) 0.0894 & (17) 0.2843 & (18) -0.3974 & (9) -0.0611 & (10) -0.3911 & (12) 11.60 & (14) 12.20 \\
\midrule
\multirow{18}{*}{\rotatebox{90}{SwinFace~\cite{qin2023swinface}}} & FaceQnet~\cite{hernandez2019faceqnet} & (18) 0.1921 & (18) -0.0345 & (18) 0.3006 & (18) -0.2406 & (18) 0.2909 & (18) -0.1534 & (14) 0.2953 & (16) -0.3651 & (18) -0.0604 & (18) -0.4562 & (15) 17.20 & (18) 17.60 \\
 & SER-FIQA~\cite{Terhorst2020SERFIQUE} & (1) \textbf{0.3858} & (1) \textbf{0.3096} & (16) 0.3170 & (16) -0.1878 & (17) 0.3444 & (17) -0.0298 & (8) 0.3066 & (6) -0.2519 & (3) -0.0148 & (2) \underline{-0.3537} & (6) 9.00 & (6) 8.40 \\
 & PFE~\cite{shi2019probabilistic} & (11) 0.2791 & (11) 0.2142 & (14) 0.3280 & (12) -0.1409 & (14) 0.3608 & (13) 0.0352 & (18) 0.2834 & (18) -0.3954 & (8) -0.0318 & (8) -0.3999 & (14) 13.00 & (17) 12.40 \\
 & MagFace~\cite{Meng2021MagFaceAU} & (10) 0.3152 & (3) 0.2975 & (7) 0.3365 & (7) -0.1148 & (10) 0.3668 & (9) 0.0503 & (11) 0.3005 & (11) -0.3213 & (19) -0.1036 & (19) -0.4664 & (11) 11.40 & (8) 9.80 \\
 & SDD-FIQA~\cite{Ou2021SDDFIQAUF} & (15) 0.2237 & (12) 0.1540 & (11) 0.3292 & (11) -0.1402 & (13) 0.3642 & (10) 0.0486 & (9) 0.3052 & (10) -0.3007 & (16) -0.0540 & (12) -0.4239 & (13) 12.80 & (11) 11.00 \\
 & LightQNet~\cite{9528058} & (7) 0.3207 & (5) 0.2893 & (15) 0.3225 & (15) -0.1674 & (15) 0.3586 & (14) 0.0315 & (16) 0.2890 & (13) -0.3326 & (12) -0.0428 & (13) -0.4303 & (14) 13.00 & (15) 12.00 \\
 & FaceQGen~\cite{HernandezOrtega2021FaceQgenSD} & (12) 0.2670 & (13) 0.1486 & (17) 0.3140 & (17) -0.1912 & (16) 0.3519 & (16) 0.0188 & (5) 0.3139 & (4) -0.2336 & (1) \textbf{-0.0059} & (3) -0.3550 & (8) 10.20 & (10) 10.60 \\
 & FaceQAN~\cite{Babnik2022FaceQANFI} & (13) 0.2313 & (14) 0.1318 & (10) 0.3307 & (9) -0.1223 & (8) 0.3707 & (7) 0.0573 & (12) 0.2995 & (12) -0.3214 & (15) -0.0510 & (14) -0.4324 & (12) 11.60 & (12) 11.20 \\
 & CR-FIQA (L)~\cite{Boutros2021CRFIQAFI} & (6) 0.3359 & (10) 0.2183 & (13) 0.3280 & (13) -0.1433 & (7) 0.3739 & (8) 0.0571 & (13) 0.2959 & (14) -0.3369 & (14) -0.0508 & (16) -0.4395 & (9) 10.60 & (16) 12.20 \\
 & DifFIQA~\cite{babnik2023diffiqa} & (14) 0.2301 & (16) 0.0744 & (4) 0.3448 & (4) -0.0736 & (6) 0.3766 & (6) 0.0643 & (2) \underline{0.3216} & (2) \underline{-0.2137} & (5) -0.0253 & (4) -0.3760 & (4) 6.20 & (5) 6.40 \\
 & DifFIQA (R)~\cite{babnik2023diffiqa} & (16) 0.2159 & (17) 0.0539 & (6) 0.3411 & (6) -0.0922 & (3) 0.3828 & (4) 0.0846 & (15) 0.2924 & (15) -0.3490 & (7) -0.0298 & (9) -0.4018 & (7) 9.40 & (9) 10.20 \\
 & CLIB-FIQA~\cite{Ou_2024_CVPR} & (4) 0.3644 & (7) 0.2664 & (2) \underline{0.3460} & (3) -0.0733 & (4) 0.3825 & (3) 0.0867 & (10) 0.3013 & (9) -0.2967 & (4) -0.0249 & (5) -0.3820 & (2) \underline{4.80} & (4) 5.40 \\
 & GraFIQs~\cite{kolf2024grafiqs} & (8) 0.3183 & (8) 0.2507 & (12) 0.3284 & (14) -0.1487 & (12) 0.3646 & (12) 0.0404 & (4) 0.3165 & (7) -0.2583 & (6) -0.0289 & (7) -0.3902 & (5) 8.40 & (7) 9.60 \\
 & eDifFIQA (S)~\cite{10468647} & (5) 0.3511 & (9) 0.2469 & (3) 0.3452 & (2) \underline{-0.0715} & (5) 0.3788 & (5) 0.0726 & (3) 0.3179 & (3) -0.2195 & (10) -0.0370 & (6) -0.3848 & (3) 5.20 & (3) 5.00 \\
 & eDifFIQA (M)~\cite{10468647} & (3) 0.3803 & (4) 0.2974 & (1) \textbf{0.3542} & (1) \textbf{-0.0413} & (2) \underline{0.3872} & (2) \underline{0.1003} & (7) 0.3088 & (5) -0.2427 & (13) -0.0492 & (11) -0.4218 & (3) 5.20 & (2) \underline{4.60} \\
 & eDifFIQA (L)~\cite{10468647} & (2) \underline{0.3826} & (2) \underline{0.3096} & (5) 0.3437 & (5) -0.0834 & (1) \textbf{0.3882} & (1) \textbf{0.1054} & (1) \textbf{0.3333} & (1) \textbf{-0.1382} & (2) \underline{-0.0120} & (1) \textbf{-0.3300} & (1) \textbf{2.20} & (1) \textbf{2.00} \\
 & FROQ~\cite{Babnik2025FROQ1OF} & (17) 0.2069 & (15) 0.1204 & (8) 0.3332 & (8) -0.1211 & (9) 0.3678 & (11) 0.0424 & (6) 0.3090 & (8) -0.2777 & (11) -0.0424 & (15) -0.4391 & (8) 10.20 & (13) 11.40 \\
 & ViT-FIQA~\cite{atzori2025vit} & (9) 0.3181 & (6) 0.2847 & (9) 0.3321 & (10) -0.1322 & (11) 0.3646 & (15) 0.0294 & (17) 0.2875 & (17) -0.3933 & (9) -0.0341 & (10) -0.4190 & (10) 11.00 & (14) 11.60 \\
\bottomrule
\end{tabular}
\end{adjustbox}
\caption{\textbf{Comparison of FIQA methods using RCE.} 
RCE scores are shown for every dataset–FR‑model–FIQA‑method combination under two weighting schemes: $\gamma = 0$ (equal weighting) and $\gamma = 3$ (greater emphasis on low‑quality samples). Ranks (lower is better) are given in parentheses. The Summary block reports the average rank of each method across all datasets for the corresponding $\gamma$ setting. Increasing $\gamma$ reorders the methods. Highlighted are the best/second-best results reveal a state-of-the-art FIQA ranking that differs markedly from those produced by EDC-based protocols.}
\label{tab:fiqa_ranking}
\vspace{-4.5mm}
\end{table*}


\section{Conclusion}

This work shows that the widely used EDC protocol for FIQA evaluation suffers from two intrinsic flaws: \textit{Test‑Set Divergence} and \textit{Threshold Drift}, which create inconsistent test sets and operating points across FIQA methods and thus undermine benchmark reliability. Extensive experiments show that the standard EDC, as well as our proposed EDC variants, remain limited. To overcome these issues, we introduce Rank Consistency Evaluation (RCE), which evaluates the entire test set with a fixed decision threshold and measures the rank‑based correlation between predicted quality and recognition errors. Extensive experiments across multiple FR models, datasets, and FIQA methods show that RCE produces markedly different rankings: recent state-of-the-art approaches that top EDC rankings perform substantially worse under RCE, revealing that EDC overestimates the effectiveness of certain solutions. We therefore recommend adopting rank-based metrics, such as RCE, for future evaluations to ensure reliability and comparability. 
Although validated only on the face images, the identified EDC limitations and proposed approaches are modality-independent, suggesting broader applicability of our framework to biometric quality assessment in general.

\paragraph{Acknowledgments -}
This work was funded by the Deutsche Forschungsgemeinschaft (DFG, German Research Foundation) under Grant 544631027.

\clearpage
\clearpage
{\small
\bibliographystyle{ieee}
\bibliography{egbib}
}

\end{document}